\documentclass[journal]{IEEEtran}

\usepackage{times}
\usepackage{epsfig}
\usepackage{graphicx}
\usepackage{amsmath,amssymb} 
\usepackage{amssymb}
\usepackage[export]{adjustbox}
\usepackage{array}
\usepackage{rotating}
\usepackage{multirow}
\usepackage{booktabs} 
\usepackage{csquotes}
\usepackage{multirow}
\usepackage{color}
\usepackage{color, colortbl}
\usepackage{float}

\makeatletter
\def\BState{\State\hskip-\ALG@thistlm}
\makeatother

\definecolor{Gray}{gray}{0.9}

\usepackage{xspace}
\newcommand{\latinphrase}[1]{\textit{#1}}  
\newcommand{\etal}{\latinphrase{et~al.}\xspace}
\newcommand{\ie}{\latinphrase{i.e.}\xspace}

\newcommand{\etc}{\latinphrase{etc.}\xspace}

\usepackage{algorithm}
\usepackage{algorithmic}

\usepackage[pagebackref=true,breaklinks=true,letterpaper=true,colorlinks,bookmarks=false]{hyperref}

\usepackage[space, nocompress]{cite}

\begin{document}
\title{Chaining Identity Mapping Modules for Image Denoising}

\author{Saeed Anwar,~\IEEEmembership{Student Member,~IEEE,}
       ~Cong Phuoc Huynh,~\IEEEmembership{Member,~IEEE and Fatih Porikli,~\IEEEmembership{Fellow,~IEEE}
       }
\thanks{S. Anwar and F. Porikli are with the Research School of Engineering, Australian National University, and CSIRO Data61. This research was done while C. P. Huynh was with the Research School of Engineering, Australian National University. This research was supported under Australian Research Council's Discovery Projects funding scheme (project number DP150104645).
\protect\\E-mail:~(saeed.anwar, fatih.porikli, cong.huynh)@anu.edu.au}}


\maketitle

\begin{abstract}
We propose to learn a fully-convolutional network model that consists of a Chain of Identity Mapping Modules (CIMM) for image denoising. The CIMM structure possesses two distinctive features that are important for the noise removal task. Firstly, each residual unit employs identity mappings as the skip connections and receives pre-activated input in order to preserve the gradient magnitude propagated in both the forward and backward directions. Secondly, by utilizing dilated kernels for the convolution layers in the residual branch, each neuron in the last convolution layer of each module can observe the full receptive field of the first layer. 

After being trained on the BSD400 dataset, the proposed network produces remarkably higher numerical accuracy and better visual image quality than the classical state-of-the-art and CNN algorithms when being evaluated on conventional benchmark images, the BSD68 dataset and real-world images from Darmstadt Noise Dataset (DND).


\end{abstract}


\IEEEpeerreviewmaketitle

\section{Introduction}

\begin{figure}[!t]
\begin{center}
\begin{tabular}{c@{ } c}  
\includegraphics[trim={0cm 3cm  0cm  1.5cm },clip,width=.23\textwidth]{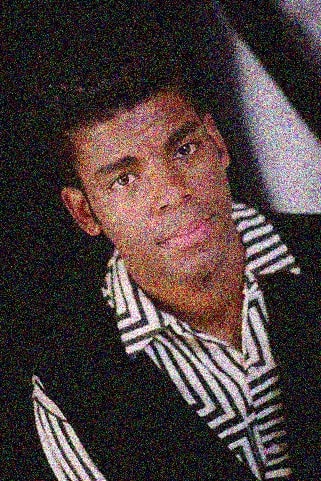} &  
\includegraphics[trim={0cm 3cm  0cm  1.5cm },clip,width=.23\textwidth]{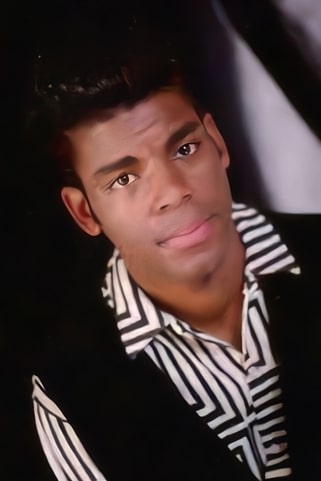}\\
 (a) Input (14.16 dB) & (b) Proposed (32.64 dB)\\
\includegraphics[trim={0cm 3cm  0cm  1.5cm },clip,width=.23\textwidth]{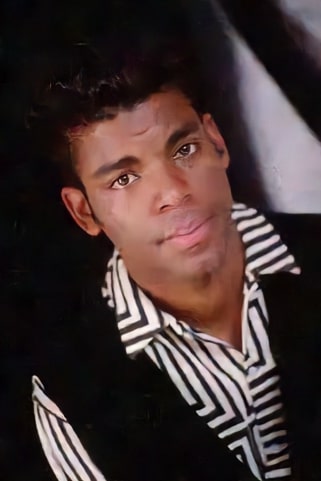} &  
\includegraphics[trim={0cm 3cm  0cm  1.5cm },clip,width=.23\textwidth]{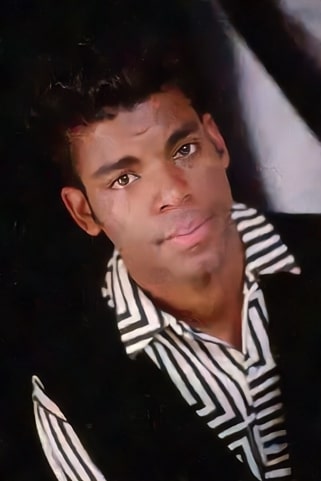}\\
 (c) IRCNN~\cite{zhang2017IRCNN} (32.07 dB) & (d) DnCNN~\cite{zhang2017DnCNN} (32.05 dB)\\
\end{tabular}
\end{center}
\vspace*{-3mm}
\caption {Denoising results for an image corrupted by the Gaussian noise with $\sigma=50$. Our result has the best PSNR score, and unlike other methods, it does not have over-smoothing or over-contrasting artifacts. Best viewed in color on high-resolution display.}
\label{fig:sample_front_image}
\vspace*{-5mm}
\end{figure}

In recent years, the amount of multimedia content is growing at an enormous rate, for example, online videos, audios, and photos due to hand-held devices and other types of multimedia devices. Thus, image processing, specifically, the image denoising has become an essential process for various computer vision and image analysis applications.  Few notable methods benefiting from image denoising on multimedia applications are detection \cite{Liu2018TMMObject}, face recognition \cite{Liu2016FaceTMM}, image Inpaiting\cite{Liu2018InpaintingTMM}, artifact removal \cite{Galteri2019ArtifactTMM}, image deblurring \cite{Yu2014Deblurring},  super-resolution \cite{Xie2015SRDenoiseTMM} and feature learning in multimedia \cite{Yang2015learningTMM}. In the past few years, the research focus in this area has been shifted to how to make the best use of image priors. To this end, several approaches attempted to exploit non-local self similar (NSS) patterns~\cite{Buades2005NLM,Dabov2007BM3D}, sparse models~\cite{Gu2014WNN,peng2012rasl,Huang2014SparseTMM}, gradient models~\cite{xu2007Iterative,weiss2007makes}, Markov random field models~\cite{roth2009fields}, external denoising~\cite{Yue2014CID,anwar2017category,luo2015adaptive} and convolutional neural networks~\cite{zhang2017DnCNN, lefkimmiatis2017NLNet, zhang2017IRCNN}.

The non-local matching (NLM) of self-similar patches and block matching with 3D filtering (BM3D) in a collaborative  manner have been two prominent baselines for image denoising for almost a decade now. 
Due to popularity of NLM~\cite{Buades2005NLM} and BM3D~\cite{Dabov2007BM3D}, a number of their variants~\cite{Foi2007SADCT,Lebrun2013NLB,Goossens2008INLM} were also proposed to execute the search for similar patches in similar transform domains.

Complementing above, use of external priors for denoising has been motivated by the pioneering studies in ~\cite{Levin2011Bounds,Chatterjee2010IDD}, which showed that selecting correct reference patches from a large external image dataset of clean samples can theoretically suppress additive noise and attain infinitesimal reconstruction error. However, directly incorporating patches from an external database is computationally prohibitive even for a single image. To overcome this problem, Chan~\etal~\cite{chan2014monte} proposed efficient sampling techniques for large databases but still the denoising is impractical as it takes hours to search patches for one single image if not days. An alternative to these methods can be considered as the dictionary learning based approaches~\cite{Elad2009ERD,Mairal2009NLSM,Dong2011CSR}, which learn over-complete dictionaries from a set of external natural clean images and then enforce patch self-similarity through sparsity. Similarly, the work in \cite{Zha2017Group} imposed a group residual representation between the sparse representation of the noise contaminated image and that of its prefiltered version to minimize the error.

Towards an efficient fusion of external datasets, many previous works~\cite{Zoran2011EPLL,Chen2015External,Xu2015PG-GMM} investigated the use of maximum likelihood frameworks to learn Gaussian mixture models of natural image patches or group patches for clean patch estimation. Several studies, including ~\cite{Xu2015PGPD,chen2015PCLR}, modified Zoran~\etal~\cite{Zoran2011EPLL}'s statistical prior for reconstruction of class-specific noisy images by capturing the statistics of noise-free patches from a large database of same category images through the Expectation-Maximization algorithm. Other similar methods on external denoising include TID~\cite{luo2015adaptive}, CSID~\cite{anwar2017category} and CID~\cite{Yue2015CID}; however, all of these have limited applicability in denoising of generic (from an unspecific class) images.

As an alternative, CSF \cite{schmidt2014CSF} learns a single framework based on unification of random-field based model and half-quadratic optimization. The role of the shrinkage in wavelet image restoration is to attenuate small values towards zero due to the assumption of these values being the product of noise instead of the signal values. The pixel values of the shrinkage mappings are learned discriminatively. These predictions are then chained to form a cascade of shrinkage fields of Gaussian conditional random Fields. The CSF algorithm considers the data term to be quadratic and must have a closed-form solution based on discrete Fourier transform.

With the rise of convolutional neural networks (CNN), a significant performance boost for image denoising has been achieved \cite{zhang2017DnCNN,zhang2017IRCNN,lefkimmiatis2017NLNet, Burger2012MLP,schmidt2014CSF}. Using deep neural networks, IrCNN \cite{zhang2017IRCNN} and DnCNN \cite{zhang2017DnCNN} learn the residual present in the contaminated image by using the noise in the loss function instead of the clean image as the ground-truth. The architectures of IrCNN \cite{zhang2017IRCNN} and DnCNN \cite{zhang2017DnCNN} are very simple as it only stacks of convolutional, batch normalization and ReLU layers. Although both models were able to report favorable results, their performance depends heavily on the accuracy of noise estimation without knowing the underlying structures and textures present in the image. Besides, they also learn batch normalization parameters after every convolutional layer.  

TRND~\cite{chen2017TNRD} incorporated a field-of-experts prior~\cite{roth2009fields} into its convolutional network by extending conventional nonlinear diffusion model to highly trainable parametrized linear filters and influence functions. It has shown improved results over more classical methods; however, the imposed image priors inherently impede its performance, which highly rely on the choice of hyper-parameter settings, extensive fine-tuning and stage-wise training.

Another notable deep learning based work is non-local color image denoising abbreviated as NLNet \cite{lefkimmiatis2017NLNet} which exploits the non-local self-similarity using deep networks. Non-local variational schemes have motivated the design of the NLNet model \cite{lefkimmiatis2017NLNet} and employ the non-local self-similarity property of natural images for denoising.  The performance heavily depends on coupling discriminative learning and self-similarity. The restoration performance is comparatively better to several earlier state-of-the-art. Though, this model improves on classical methods but lagging behind  IrCNN~\cite{zhang2017IRCNN} and DnCNN~\cite{zhang2017DnCNN}, as it inherits the limitations associated with the NSS priors as not all patches recur in an image.

FormResNet is proposed by \cite{jiao2017formresnet} which builds  upon DnCNN \cite{zhang2017DnCNN}. This model is composed of two networks; both networks are similar to DnCNN \cite{zhang2017DnCNN}; however, the difference lies in the loss layers. The first network termed as ``Formatting layer''  has incorporated the Euclidean and perceptual loss into one. The Classical algorithms such as BM3D can also replace this formatting layer. Although the second deep network is exactly similar to DnCNN \cite{zhang2017DnCNN}; however, the authors named it ``DiffResNet'' and input to this network is fed from the first one. The stated formatting layer removes high-frequency corruption in uniform areas, while DiffResNet learns the structured regions. FormResNet \cite{jiao2017formresnet} improves upon the results of DnCNN~\cite{zhang2017DnCNN} by a small margin. 

Recently, wavelet domain deep network denoising architecture for CNN is proposed by \cite{bae2017beyond}. The motivation behind this CNN based network is persistent homology analysis \cite{edelsbrunner2008persistent}. The network takes wavelet transformed images as input and learns the features in transformed manifold rather than original image manifold. The proposed network has high number of channels as compared to DnCNN \cite{zhang2017DnCNN} and FormResNet \cite{jiao2017formresnet}; hence, the marginal increase in PSNR can be attributed to the number of channels employed for learning features.

\subsection{Inspiration \& Motivation}


\begin{figure}
\begin{center}
\includegraphics[width=0.45\textwidth]{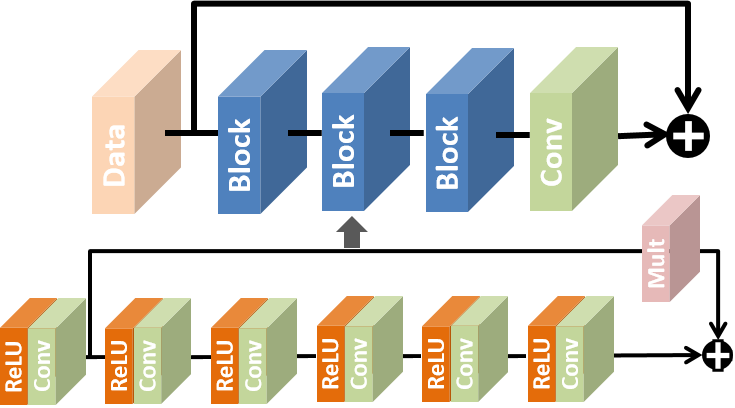}\\ 
\end{center}
\vspace*{-3mm}
\caption {The proposed network architecture, which consists of multiple modules with similar structures. Each module is composed of a series of pre-activation-convolution layer pairs.}
\label{fig:net_architecture}
\vspace*{-5mm}
\end{figure}

Existing convolutional neural network based image denoising methods~\cite{Burger2012MLP,zhang2017DnCNN,zhang2017IRCNN} connect weight layers consecutively and learn the mapping by brute force. One problem with such an architecture is the addition of more weight layers to increase the depth of the network. Even if the new weight layers are added to the mentioned CNN based denoising methods, it will fall into gradients vanishing problem and impel it further~\cite{bengio1994vanishing}. This property of increasing the size of the network is important and helps in performance boost~\cite{he2016deep}. Therefore, our goal is to propose a model that overcomes this deficiency. 

Another reason is the lack of true color denoising. Most of the current denoising systems are either for grayscale image denoising or treat each color channel separately ignoring the relationship between the color channels. Only a handful of works~\cite{dabov2007CBM3D,anwar2017category,zhang2017DnCNN,lefkimmiatis2017NLNet} approached color image denoising in its own context. 

To provide a solution, our choice is the convolutional neural networks in a discriminative prior setting for image denoising. There are many advantages of using CNNs, including efficient inference,  incorporation of robust priors, integration of local and global receptive fields, regressing on nonlinear models, and discriminative learning capability. Furthermore, we propose a modular network where we call each module as a mapping modules (MM). The mapping modules can be replicated and easily extended to any arbitrary depth for performance enhancement. 


\subsection{Contributions}
The contributions of this work can be summarized as follows:
\begin{itemize}
\item An effective CNN architecture that consists of a Chain of Identity Mapping modules (CIMM) for image denoising. These modules share a common composition of layers, with residual connections between them to facilitate training stability.
\item The use of dilated convolutions for learning suitable filters to denoise at different levels of spatial extent. 
\item A single denoising network that can handle various noise levels.
\end{itemize}

\section{Chain of Identity Mapping Modules}
\label{sec:CIMM}

This section presents our approach to image denoising by learning a Convolutional Neural Network consisting of a Chain of Identity Mapping Modules (CIMM). Each module is composed of a series of pre-activation units followed by convolution functions, with residual connections between them. 
Section~\ref{sec:learning_obj} formulates the learning objective. Subsequently, the meta-structure of the CIMM network in Section~\ref{sec:network_architecture}. 

\subsection{Network Design}
\label{sec:network_architecture}
Residual learning has recently delivered state of the art results for object classification~\cite{He15Residual,He2016IM} and detection~\cite{Lin16FPN}, while offers training stability. Inspired by the Residual Network variant with identity mapping~\cite{He2016IM}, we adopt a modular design for our denoising network. The design consists of a Chain of Identity Mapping modules (CIMM). 

\subsubsection{Network elements}

Figure~\ref{fig:net_architecture} depicts the entire architecture, where identity mapping modules are shown as a blue blocks, which are in turn composed of basic ReLU (orange) and convolution (green) layers. The output of each module is a summation of the identity function and the residual function. In our experiments, we typically employ $64$ filters of size $3\times3$ in each convolution layer.

The meta-level structure of the network is governed by three parameters: the number of identity modules (\ie $\mathbf{M}$), the number of pre-activation-convolution pairs in each module (\ie $\mathbf{L}$), and the number of output channels (\ie $\mathbf{C}$), which we fixed across all the convolution layers.

The high-level structure of the network can be viewed as a chain of identity mapping modules, where the output of each module is fed directly into the subsequent one. Subsequently, the output of this chain is fed to a final convolution layer to produce a tensor with the same number of channels as the input image. At this point, the final convolution layer directly predicts the noise component from a noisy image. The noise-free image/patch is then subtracted from the input to recover the noise-free image.

The identity mapping modules are the building blocks of the network, which share the following structure. Each module consists of two branches: a residual branch and an identity mapping branch. The residual branch of each module contains a series of layers pairs, \ie a nonlinear pre-activation (typically ReLU) layer, followed by a convolution layer. Its main responsibility is to learn a set of convolution filters to predict image noise. In addition, the identity mapping branch in each module allows the propagation of the loss gradients in both directions without any bottleneck. 

\subsubsection{Justification of design}
For image denoising, several previous works have adopted a fully convolutional network design, without any pooling mechanism~\cite{zhang2017DnCNN,kim2016VDSR,zhang2017IRCNN}. This is necessary in order to preserve the spatial resolution of the input tensor across different layers. We follow this design by using only non-linear activations and convolution layers across our network.

Furthermore, we aim to design the network in such a way where convolution layers neurons in the last layer of each identity mapping (IM)  module observe the full spatial receptive field in the first convolution layer. This design helps learning to connect input neurons at all spatial locations to the output neurons, in much the same way as well-known non-local mean methods such as~\cite{Dabov2007BM3D,Buades2005NLM}. Instead of using a unit stride within each layer, we also experimented with dilated convolutions to increase the receptive fields of the convolution layers. By this design, we can reduce the depth of each IM module while the final layer's neurons can still observe the full input spatial extent.



Pre-activation has been shown to offer the highest performance for classification when used together with identity mapping~\cite{He2016IM}. In a similar fashion, our design employs ReLU before each convolution layer. This design differs from existing neural network architectures for denosing~\cite{kim2016VDSR,zhang2017DnCNN,lefkimmiatis2017NLNet}. The pre-activation helps training to converge more easily, by while the identity function preserves the range of gradient magnitudes. Also, the resulting network generalizes better as compared to the post-activation alternative. This property enhances the denoising ability of our network.

\subsubsection{Formulation}
Now we formulate the prediction output of this network structure for a given input patch $\textbf{y}$. Let $\mathcal{W}$ denote the set of all the network parameters, which consists of the weights and biases of all constituting convolution layers. Specifically, we let $w_{m, l}$ denote both the kernel and bias parameters of the $l$-th convolution layer in the residual branch of the $m$-th module. 
 
Within such a branch, the intermediate output of the $l$-th ReLU-convolution pair and of the $m$-th module is a composition of two functions
\begin{equation}
\mathbf{z}_{m, l} = f(g(\mathbf{y}_{m, l}); w_{m, l}),
\end{equation}
where $f$ and $g$ are the notation for the convolution and the ReLU functions,  $\mathbf{z}_{m, l}$ is the output of the $l$-th ReLU-convolution pair of $m$-th module. 
By composing the series of ReLU-convolution pairs, we obtain the output of the $m$-th residual branch as
\begin{equation}
\begin{split}
\mathbf{r}_{m} = -\mathbf{z}_{m, 0} + f(g( \ldots f(g(\mathbf{y}_{m,0}; w_{m, 0})) \ldots ); w_{m, l})
\end{split}
\end{equation}

where $\mathbf{z}_{m, 0}$ is the output of the first ReLU-convolution pair. Chaining all the identity mapping modules, we obtain the output as $\sum_{m=1}^{M} \textbf{r}_m$. Finally, the output of this chain is convolved with a final convolution layer with learnable parameters $w_{m+1}$ to  predict the noise component as 
$h(\mathbf{y}, \mathcal{W}) = f(\mathbf{y} + \sum_{m=1}^{M} \textbf{r}_m, w_{m+1})$.

\subsection{Learning to Denoise}
\label{sec:learning_obj}

Our convolutional neural network (CNN) is trained on image patches or regions rather than at the image-level. This decision is driven by a number of reasons. 
Firstly, it offers random sampling of a large number of training samples at different locations from various images. Random shuffling of training samples is well-known to be a useful technique to stabilize the training of deep neural networks. 
Therefore, it is preferable to batch training patches with a random, diverse mixture of local structures, patterns, shapes and colors.
Secondly, there has been success in approaches that learns image patch priors from external data for image denoising~\cite{Zoran2011EPLL}.

From a set of noise-free training images, we randomly crop a number of training patches $\mathbf{x}_i, i=i, \ldots, N$ as the groundtruth. The noisy version of these patches is obtained by adding (Gaussian) noise to the ground truth training images. Let us denote the set of noisy patches corresponding to the former as $\mathbf{y}_i, i=i, \ldots, N$. With this setup, our image denoising network (described in Section~\ref{sec:network_architecture}) is aimed to reconstruct a patch $\mathbf{x}^*_i =h(\mathbf{y}_i, \mathcal{W})$ from the input patch $\mathbf{y}_i$. 

The learning objective is to minimize the following sum of squares of $\ell_2$-norms 
\begin{equation}
\mathcal{L} \triangleq \frac{1}{N}\sum_{i=1}^{N}  \Vert h(\mathbf{y}_i, \mathcal{W}) - \mathbf{x}_i \Vert^2
\label{eq:objective}
\end{equation}

To train the proposed network on a large dataset, we minimization of the objective function in Equation~\ref{eq:objective} on mini-batches of training examples. Training details for our experiments are described in Section \ref{label:training}.

\begin{table}[!t]
\caption{Denoising performance (in PSNR) on the BSD68 dataset~\cite{Martin2001BSD} for different sizes of training input patches for $\sigma_n = 25$, keeping all other parameters constant.}
\centering
\begin{tabular}{|c|c|c|c|c|c|}
\cline{1-6}
\multicolumn{6}{|c|}{Training patch size}\\ \hline
  20 	  & 30  	 & 40 		& 50 		&60  & 70\\ \hline 
  29.13  & 29.30 	 & 29.34   & 29.36    & 29.37 & 29.38\\ 
\hline
\end{tabular}
\label{table:Training_patch_size}
\vspace*{-3mm}
\end{table}

\begin{table}
\caption{The average PSNR of the denoised images for the BSD68 dataset, with respect to different number of modules $\textbf{M}$. The higher the number of modules, the higher is the accuracy.}
\centering
\begin{tabular}{|c|c|c|c|}
\cline{1-4}
\multicolumn{4}{|c|}{Number of modules}                \\ \hline

2   	  &4    	   		&6 				&8   \\ \hline 
29.28	  &29.34 	&29.35  	& 29.36	 \\ \hline
\end{tabular}
\label{table:No_of_cnn_modules}
\vspace*{-5mm}
\end{table}

\section{Experiments}


\subsection{Datasets and Baselines}
\label{sec:Datasets}
We performed experimental validation on the widely used classical images (same number and images as \cite{zhang2017DnCNN}). Similarly, we also use DnD datasets \cite{plotz2017benchmarking} consists of real 1000 images and BSD68 dataset \cite{roth2009fields} composed of 68 images. It is to be noted here, that our BSD400 dataset \cite{Martin2001BSD} for training and BSD68 dataset \cite{roth2009fields} for testing are disjoint. To generate noisy test images, we corrupt the  images by additive white Gaussian noise with standard deviations (std) of $\sigma_n=15, 25, 50, 70$, as employed by~\cite{zhang2017IRCNN,zhang2017DnCNN,lefkimmiatis2017NLNet}. For evaluation purposes, we use the Peak Signal-to-Noise Ratio (PSNR) index as the error metric. We compare our proposed method with numerous state-of-the-art methods, including BM3D~\cite{Dabov2007BM3D}, WNNM~\cite{Gu2014WNN}, MLP~\cite{Burger2012MLP}, EPLL~\cite{Zoran2011EPLL}, TNRD~\cite{chen2017TNRD}, IRCNN~\cite{zhang2017IRCNN}, DnCNN~\cite{zhang2017DnCNN} and NLNET~\cite{lefkimmiatis2017NLNet}. To ensure a fair comparison, we use the default setting provided by the respective authors.


\subsection{Training Details}
\label{label:training}
The training input to our network is noisy and noise-free patch pairs of size $40\times40$ cropped randomly from the BSD400 dataset \cite{Martin2001BSD}. Note that there is no overlap between the training and evaluation datasets. We also augment the training data with horizontally and vertically flipped versions of the original patches and those rotated at an angle of $\frac{\pi n}{2}$, where $n=1,2,3$. The training patches are randomly cropped on the fly from the 400 images of BSD400 dataset. 

We offer two strategies for handling different noise levels. The first one is to train a network for each specific noise level and we call model as noise-specific model. Alternatively, we train a single model for the noise range $[1, 50]$ (similar to ~\cite{zhang2017DnCNN}) and we refer to this model as noise-agnostic model. At each update of training, we construct a batch by randomly selecting noisy patches with noise levels between $1$ and $50$.  

We implement the denoising method in the Caffe framework on two Tesla P100 GPUs, and employ the Adam optimization algorithm~\cite{KingmaB14} for training.
The initial learning rate was set to $10^{-4}$ and the momentum parameter was $0.9$. We scheduled the learning rate such that it is halved after every ten epochs.  Our network takes six hours and 40 epochs to train a model. The input to our network are patches of $40 \times 40$ of mini-batches of size $64$. We train our network from scratch by a random initialization of the convolution weights according to the method in~\cite{He15Rectifiers} and a regularization strength, \ie weight decay, of $0.0001$.

\begin{table}
\caption{Denoising performance for different network settings to dissect the relationship between kernel dilation, number of layers and receptive field.}
\centering
\begin{tabular}{|c|c|c|c|}
\cline{1-4}
No of layers     & 18   	& 9  	    & 6      	\\ \hline
Kernel dilation  & 1   		& 2   	    & 3      	\\ \hline 
      	  		 & 29.34      & 29.34  	& 29.34  	\\ 
\hline
\end{tabular}
\label{table:layers_kernel_dilation_patchsize}
\vspace*{-5mm}
\end{table}

\subsection{Boosting Denoising Performance}
\label{sec:self_ensemble}
To boost the performance of the trained model, we use the late fusion/geometric transform strategy as adopted by~\cite{timofte2016seven}. During the evaluation, we perform eight types of augmentation (including identity) of the input noisy images $y$ as $y_i^t = \Gamma_i(y)$ where $i=1,\cdots,8$. From these geometrically transformed images, we estimate corresponding denoised images $\{\hat{x}_1^t, \hat{x}_2^t,\cdots, \hat{x}_8^t\}$, where $\hat{x}_i^t = h(\hat{y}_i^t,W)$  using our model. To generate the final denoised image $\hat{x}$, we perform the corresponding inverse geometric transform $\tilde{x}_i^{-t} = \Gamma_i^{-1}(\tilde{x}_i^t)$ and then take the average of the outputs as $\tilde{x} = \frac{1}{8}  \sum_{i=1}^{8} \tilde{x}_i^t $.
This strategy is beneficial as it saves training time and have small number of parameters as compared to individually trained eight models. We also found empirically that this fusion method gives approximately the same performance as the models trained individually with geometric transform.

\subsection{Identity Mapping Modules}
The structure of the mapping modules used in our experiments is depicted in Table~\ref{table:Mapping_Module_architecture}. Each module consists of a series of ReLU + Conv pair. All the convolution layers have a kernel size of $3\times3$ and $64$  output channels. The kernel dilation and padding are same in each layer and vary between $1$ and $3$. The skip connection connects the output of the first pair of ReLU + Conv to the last pair ReLU + Conv as shown in figure~\ref{fig:net_architecture}

\begin{table}[!t]
\caption{Detailed architecture of an identity mapping module.}
\centering
\begin{tabular}{|c|c|c|c|c|c|c|}
\hline
&\multicolumn{6}{c|}{Mapping Module Layers}\\ \cline{2-7}
Parameters  & 1$^{st}$  & 2$^{nd}$ & 3$^{rd}$ & 4$^{th}$ & 5$^{th}$ & 6$^{th}$\\ \hline 
Padding     & 1   		& 3 	   & 3        & 3        & 3        & 3\\
Dilation    & 1   		& 3 	   & 3        & 3        & 3        & 3\\ 
Kernel Size & 3   		& 3 	   & 3        & 3        & 3        & 3\\
Channels    & 64   		& 64 	   & 64       & 64       & 64       & 64\\ \hline
\end{tabular}
\label{table:Mapping_Module_architecture}
\vspace*{-4mm}
\end{table}
\subsection{Ablation Studies}

\begin{figure*}[!htbp]
\begin{center}
\begin{tabular}[b]{c@{ } c@{ }  c@{ } c@{ } c@{ } c@{ }	c@{ }}
    
    \multirow{5}{*}{\includegraphics[width=.314\textwidth,valign=t]{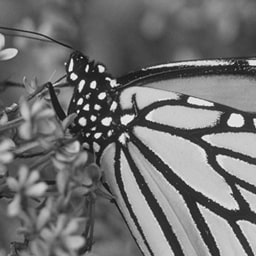}} &  
    \includegraphics[trim={3cm 3cm  3cm  3cm },clip,width=.133\textwidth,valign=t]{images/05}&
  	\includegraphics[trim={3cm 3cm  3cm  3cm },clip,width=.13\textwidth,valign=t]{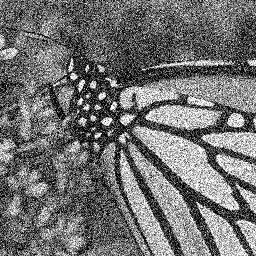}&   
    \includegraphics[trim={3cm 3cm  3cm  3cm },clip,width=.133\textwidth,valign=t]{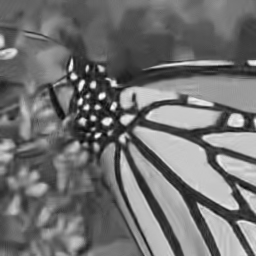}&   
	\includegraphics[trim={3cm 3cm  3cm  3cm },clip,width=.133\textwidth,valign=t]{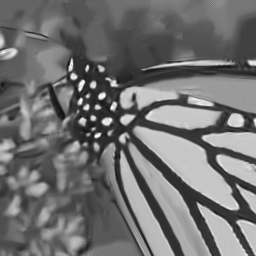}&    
  	\includegraphics[trim={3cm 3cm  3cm  3cm },clip,width=.133\textwidth,valign=t]{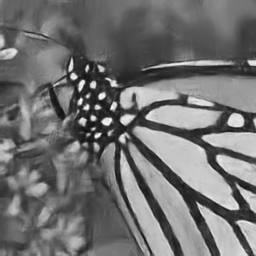}\\
    & Original& Noisy  & BM3D    & WNNM        & MLP\\
    &        &14.16dB  & 25.82dB & 26.32dB  & 26.26dB \\

    &
     \includegraphics[trim={3cm 3cm  3cm  3cm },clip,width=.133\textwidth,valign=t]{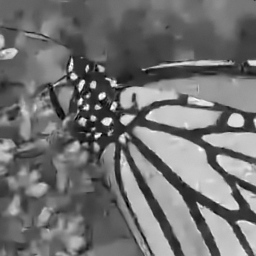}&
    \includegraphics[trim={3cm 3cm  3cm  3cm },clip,width=.133\textwidth,valign=t]{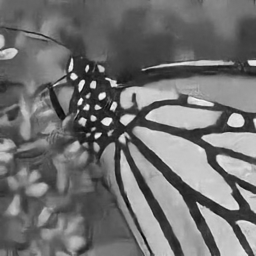}&  
    \includegraphics[trim={3cm 3cm  3cm  3cm },clip,width=.133\textwidth,valign=t]{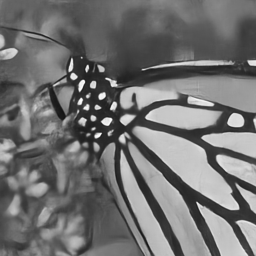}&    
  	\includegraphics[trim={3cm 3cm  3cm  3cm },clip,width=.133\textwidth,valign=t]{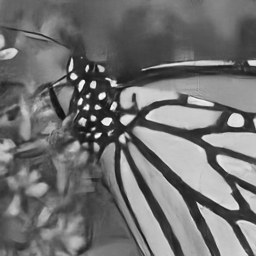}&
     \includegraphics[trim={3cm 3cm  3cm  3cm },clip,width=.133\textwidth,valign=t]{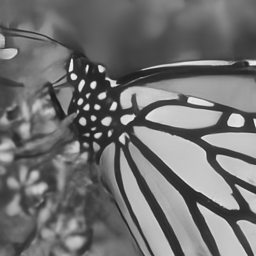}\\
      &EPLL     &TNRD     & DnCNN-S  & irCNN    & \textbf{Ours} \\
     Monarch image& 25.94dB &26.42dB & 26.78dB  & 26.61dB &       \textbf{27.21dB}  \\
    
\end{tabular}
\end{center}
\vspace*{-4mm}
\caption{Denoising quality comparison on a sample image with strong edges and texture, selected from classical image set for noise level $\sigma_n=50$. The visual quality, \ie sharpness of the edges on the wings and small textures reproduced by our method is the best among all.}
\label{fig:classical_grayscale_image}
\vspace*{-5mm}
\end{figure*}

\begin{figure*}[!htbp]
\begin{center}
\begin{tabular}{c@{ }  c@{ } c@{ } c@{ } c@{ }	c}
    \multirow{5}{*}{\includegraphics[width=.2125\textwidth,,valign=t]{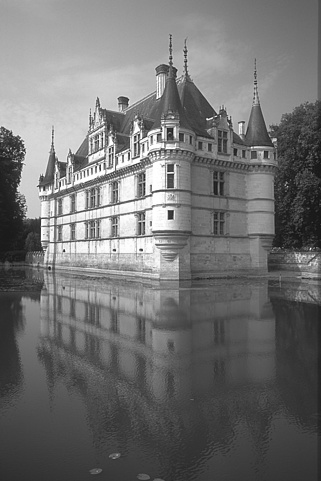}}&
    \includegraphics[trim={2cm 9.5cm  6cm  4.5cm },clip,width=.15\textwidth,valign=t]{images/test003}&  
    \includegraphics[trim={2cm 9.5cm  6cm  4.5cm },clip,width=.15\textwidth,valign=t]{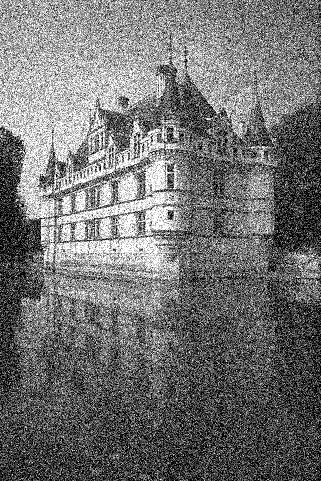}&
    \includegraphics[trim={2cm 9.5cm  6cm  4.5cm },clip,width=.15\textwidth,valign=t]{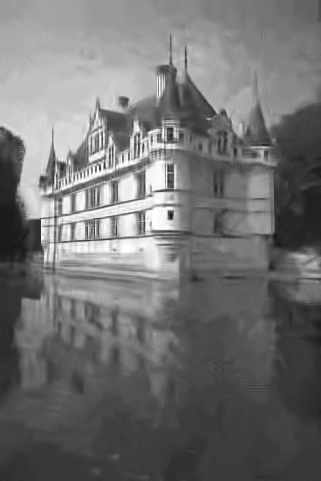}&   
	\includegraphics[trim={2cm 9.5cm  6cm  4.5cm },clip,width=.15\textwidth,valign=t]{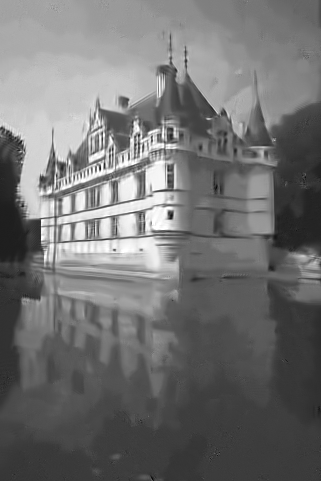}& 
  	\includegraphics[trim={2cm 9.5cm  6cm  4.5cm },clip,width=.15\textwidth,valign=t]{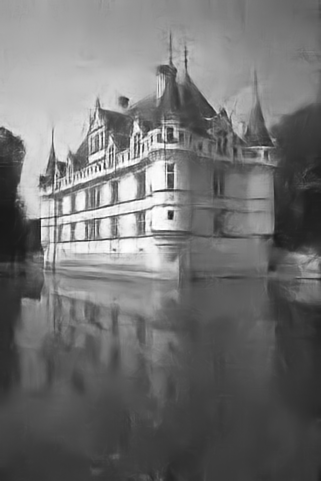}\\
    &Original& Noisy  & BM3D    & WNNM    & MLP\\
    &        &14.16dB & 26.21dB & 26.51dB & 26.54dB \\

    &
     \includegraphics[trim={2cm 9.5cm  6cm  4.5cm },clip,width=.15\textwidth,valign=t]{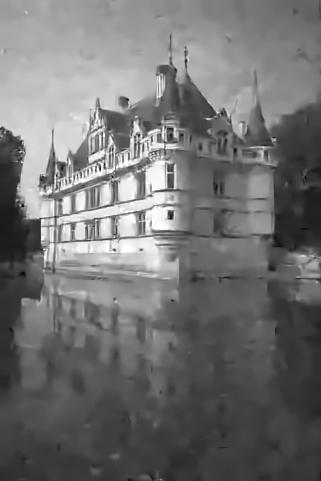}&
    \includegraphics[trim={2cm 9.5cm  6cm  4.5cm },clip,width=.15\textwidth,valign=t]{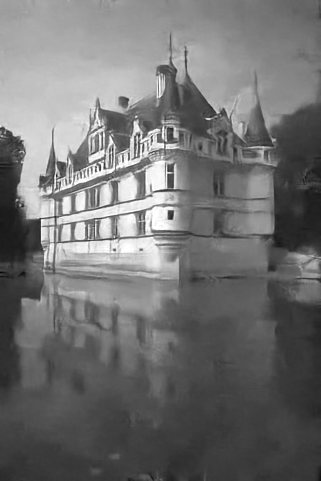}&  
    \includegraphics[trim={2cm 9.5cm  6cm  4.5cm },clip,width=.15\textwidth,valign=t]{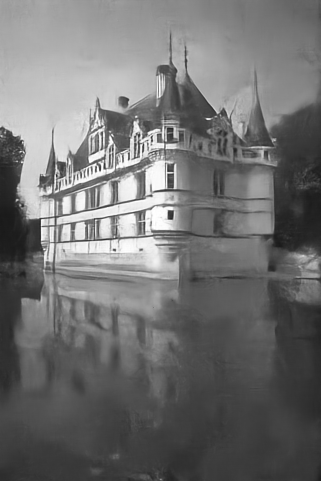}&    
  	\includegraphics[trim={2cm 9.5cm  6cm  4.5cm },clip,width=.15\textwidth,valign=t]{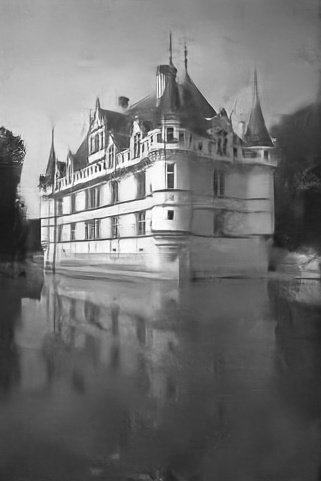}&
 
  	\includegraphics[trim={2cm 9.5cm  6cm  4.5cm },clip,width=.15\textwidth,valign=t]{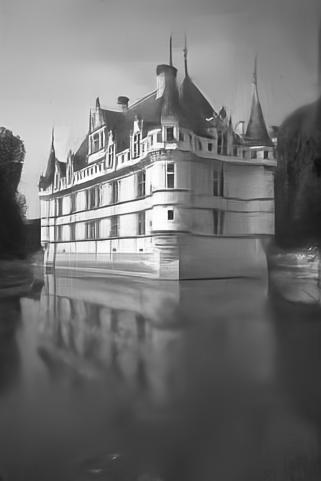}\\
     & EPLL &TNRD    & DnCNN-S & irCNN    & \textbf{Ours} \\
    Castle from BSD68~\cite{Martin2001BSD}& 26.35dB &26.60dB & 26.90dB & 26.88dB  & \textbf{27.20dB}  \\
\end{tabular}
\end{center}
\vspace*{-3.5mm}
\caption{Comparison on a sample image from BSD68 dataset~\cite{Martin2001BSD} for $\sigma_n=50$. Our network is able to recover fine textures on the castle} 
\label{fig:BSD68_grayscale_image}
\vspace*{-5mm}
\end{figure*}

\begin{figure*}[!htbp]
\begin{center}
\begin{tabular}{c@{ }  c@{ } c@{ } c@{ } c@{ }	c}
    \multirow{5}{*}{\includegraphics[width=.19\textwidth,valign=t]{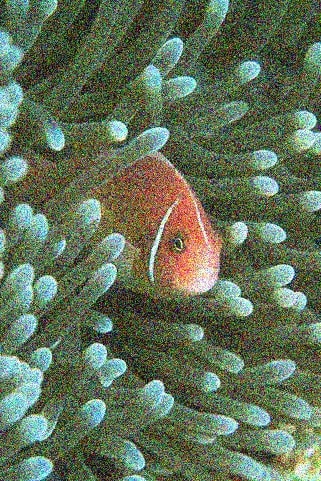}}&
    \includegraphics[trim={1.5cm 5.15cm  1.5cm  5cm},clip,width=.145\textwidth,valign=t]{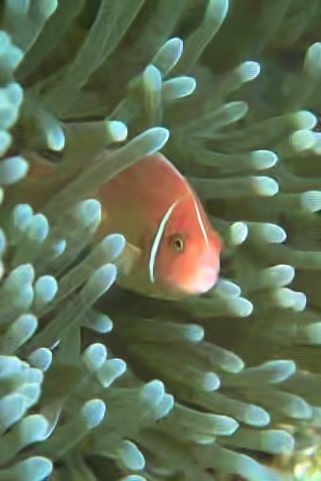}&  
    \includegraphics[trim={1.5cm 5.15cm  1.5cm  5cm},clip,width=.145\textwidth,valign=t]{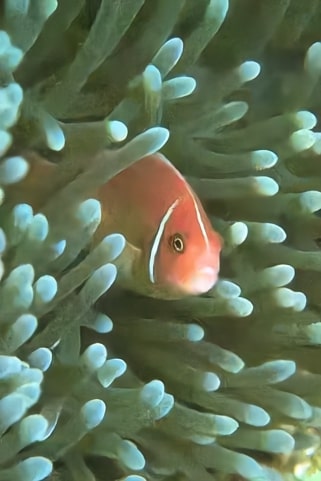}&
    \multirow{5}{*}{\includegraphics[width=.19\textwidth,valign=t]{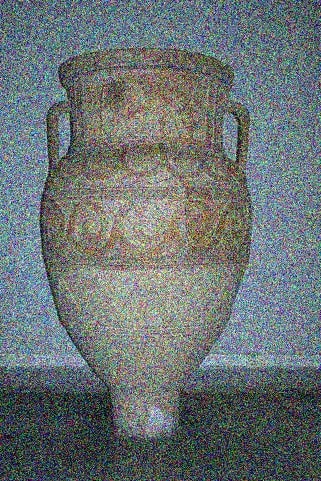}}&   
	\includegraphics[trim={1.5cm 5.15cm  1.5cm  5cm },clip,width=.145\textwidth,valign=t]{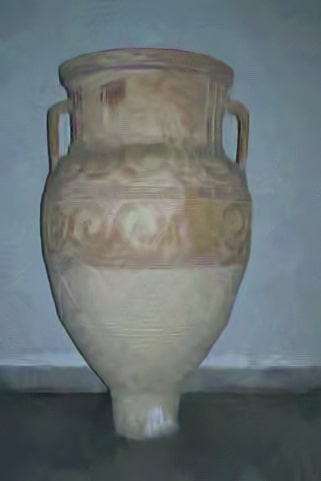}& 
  	\includegraphics[trim={1.5cm 5.15cm  1.5cm  5cm },clip,width=.145\textwidth,valign=t]{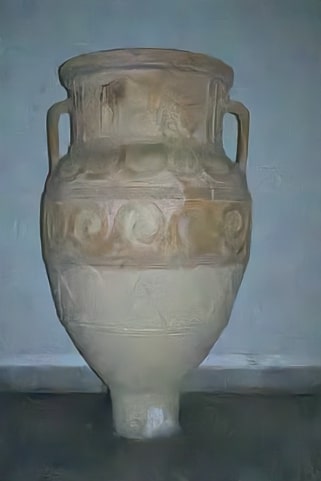}\\
 
 	&CBM3D  & DnCNN  &     & CBM3D    & DnCNN\\
    &29.65dB  &30.52dB   &     & 31.68dB    & 32.33dB  \\

     &
    \includegraphics[trim={1.5cm 5.15cm  1.5cm  5cm },clip,width=.145\textwidth,valign=t]{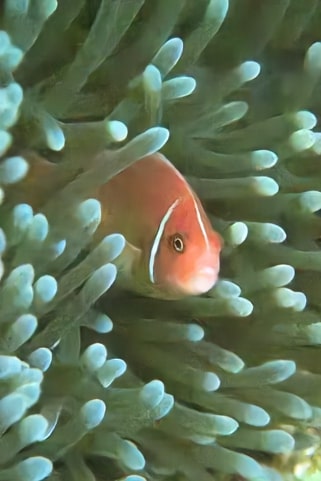}&
    \includegraphics[trim={1.5cm 5.15cm  1.5cm  5cm },clip,width=.145\textwidth,valign=t]{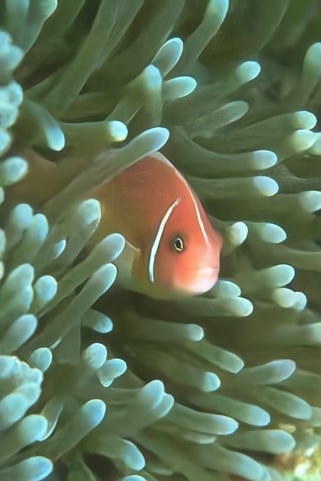}&  
    & 
    \includegraphics[trim={1.5cm 5.15cm  1.5cm  5cm},clip,width=.145\textwidth,valign=t]{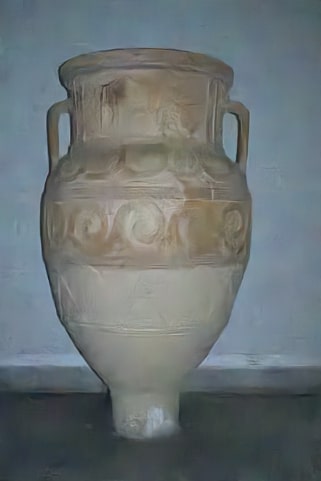}&
   	\includegraphics[trim={1.5cm 5.15cm  1.5cm  5cm },clip,width=.145\textwidth,valign=t]{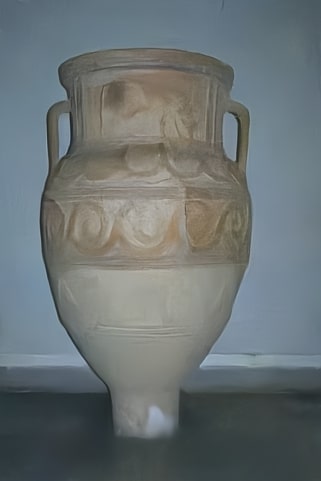}\\
 
   & irCNN &\textbf{Ours}    &  & irCNN    & \textbf{Ours} \\
    Fish from BSD68~\cite{Martin2001BSD}& 30.40dB & \textbf{31.23dB}
 &Vase from BSD68~\cite{Martin2001BSD} & 32.21dB   & \textbf{32.76dB}  \\
\end{tabular}
\end{center}
\vspace*{-3.5mm}
\caption{Denoising performance for state-of-the-art versus the proposed method on sample color images from the dataset in~\cite{Martin2001BSD}, where the noise standard deviation $\sigma_n$ is $50$. The image we recover is more natural, contains less contrast artifacts and is closest to the ground-truth.}
\label{fig:color_results_sig15}
\vspace*{-5mm}
\end{figure*}


\begin{figure*}[!htbp]
\begin{center}
\begin{tabular}{c@{} c@{} c@{} c@{}  c@{} c }

\includegraphics[trim={0cm 0cm  2cm  0cm },clip,width=.205\textwidth,valign=t]{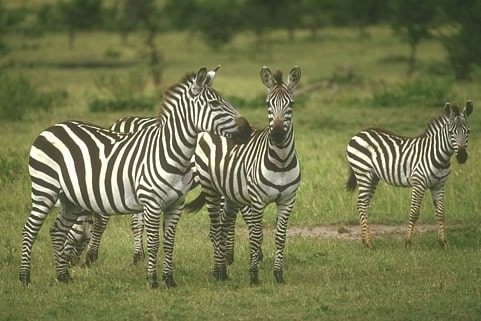}&
\includegraphics[trim={6cm 4cm  7.6cm  4cm },clip,width=.158\textwidth,valign=t]{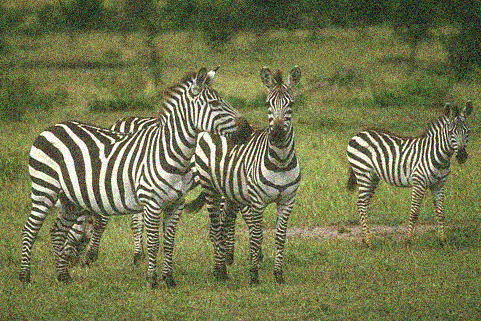}&
\includegraphics[trim={6cm 4cm  7.6cm  4cm },clip,width=.158\textwidth,valign=t]{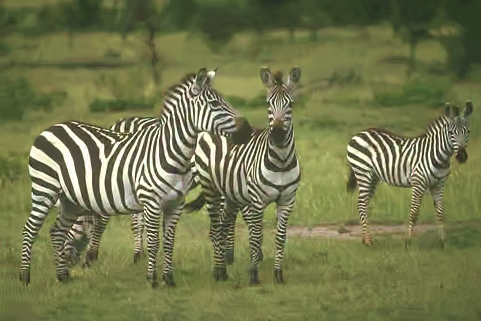}&
\includegraphics[trim={6cm 4cm  7.6cm  4cm },clip,width=.158\textwidth,valign=t]{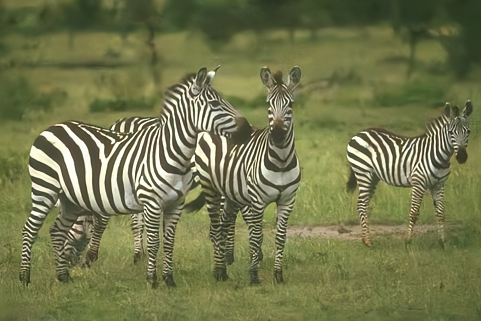}&
\includegraphics[trim={6cm 4cm  7.6cm  4cm },clip,width=.158\textwidth,valign=t]{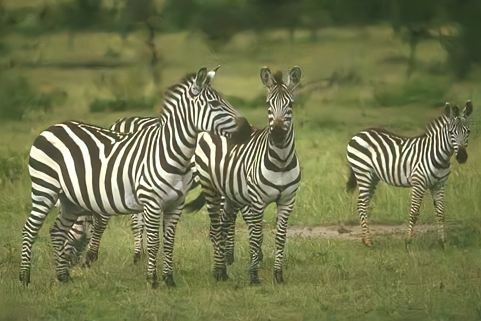}&
\includegraphics[trim={6cm 4cm  7.6cm  4cm },clip,width=.158\textwidth,valign=t]{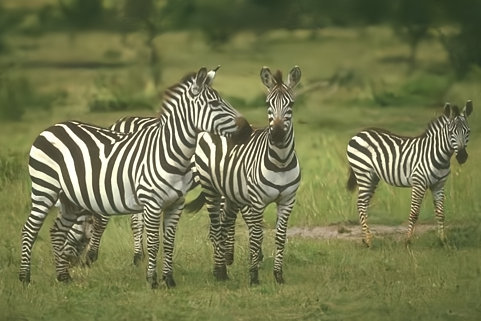}\\
Original & Input (20.18  dB)     & CBM3D (29.37 dB) & DnCNN (30.89 dB) & irCNN (30.60  dB) & Ours (\textbf{31.04} dB) \\
\end{tabular}
\end{center}
\vspace*{-3.5mm}
\caption{A sample color image with rich textures, selected from the BSD68 dataset\cite{Martin2001BSD} for $\sigma_n = 25$.}
\label{fig:color_results_sig25}
\vspace*{-5mm}
\end{figure*}

\subsubsection{Influence of the patch size} 
In this section, we show the role of the patch size and its influence on the denoising performance.
Table~\ref{table:Training_patch_size} shows the average PSNR on BSD68~\cite{roth2009fields} for $\sigma_n=25$ with respect to the increase in size of the training patch. It is obvious that there is a marginal improvement in PSNR as the patch size increases. The main reason for this phenomenon is the size of the receptive field, with a larger patch size network learns more contextual information, hence able to predict local details better.

\begin{table}
\caption{PSNR reported on the BSD68 dataset for $\sigma_n = 25$ when different features are added to the baseline (first row).}
\centering
\begin{tabular}{|c|c|c|c|c|}\hline
Dilation & Identity & Boosting & PSNR\\ 
\hline
 	 No  & No  & No  & 29.24 \\ 
	 Yes & No  & No  & 29.23 \\
	 No  & Yes & No  & 29.28 \\
	 Yes & Yes & No  & 29.32 \\
	 Yes & Yes & Yes & 29.34 \\
\hline
\end{tabular}
\label{table:feature_contributions}
\vspace*{-5mm}
\end{table}

\begin{table*}[!t]
\caption{Performance comparison between image denoising algorithms on widely used classical images, in terms of PSNR (in dB). The best results are highlighted with \textcolor{red}{red} color while the \textcolor{blue}{blue} color represents the second best denoising results.}
\begin{center}
\begin{tabular}{lccccccccccccc}
\toprule

 &\multicolumn{1}{c} {Cman}
& \multicolumn{1}{c} {House}
& \multicolumn{1}{c} {Peppers}
& \multicolumn{1}{c} {Starfish}
& \multicolumn{1}{c} {Monar}
& \multicolumn{1}{c} {Airpl}
& \multicolumn{1}{c} {Parrot}
& \multicolumn{1}{c} {Lena}
& \multicolumn{1}{c} {Barbara}
& \multicolumn{1}{c} {Boat}
& \multicolumn{1}{c} {Man}
& \multicolumn{1}{c} {Couple}
& \multicolumn{1}{c} {Average}\\ 
\toprule
& \multicolumn{13}{c}{$\sigma_n = 15$} \\ 
\cmidrule(l){1-14} 
BM3D~\cite{Dabov2007BM3D}      & 31.91 & 34.93 & 32.69 & 31.14 & 31.85 & 31.07 & 31.37 & 34.26 & \textcolor{blue}{33.10} & 32.13 & 31.92 & 32.10 & 32.372\\
WNNM~\cite{Gu2014WNN}          & 32.17 & \textcolor{blue}{35.13} & 32.99 & 31.82 & 32.71 & 31.39 & 31.62 & 34.27 & \textcolor{red}{33.60} & 32.27 & 32.11 & 32.17 & 32.696\\
EPLL~\cite{Zoran2011EPLL}      & 31.85 & 34.17 & 32.64 & 31.13 & 32.10 & 31.19 & 31.42 & 33.92 & 31.38 & 31.93 & 32.00 & 31.93 & 32.138\\
CSF~\cite{schmidt2014CSF}	   & 31.95 & 34.39 & 32.85 & 31.55 & 32.33 & 31.33 & 31.37 & 34.06 & 31.92 & 32.01 & 32.08 & 31.98 & 32.318\\
TNRD~\cite{chen2017TNRD}       & 32.19 & 34.53 & 33.04 & 31.75 & 32.56 & 31.46 & 31.63 & 34.24 & 32.13 & 32.14 & 32.23 & 32.11 & 32.502\\
DnCNNS~\cite{zhang2017DnCNN}   & \textcolor{blue}{32.61} & 34.97 & \textcolor{red}{33.30} & 32.20 & \textcolor{blue}{33.09} & \textcolor{blue}{31.70} & \textcolor{blue}{31.83} & 34.62 & 32.64 & 32.42 & \textcolor{blue}{32.46} &  32.47 & \textcolor{blue}{32.859}\\
DnCNNB~\cite{zhang2017DnCNN}   & 32.10 & 34.93 & 33.15 & 32.02 & 32.94 & 31.56 & 31.63 & 34.56 & 32.09 & 32.35 & 32.41 & 32.41 & 32.680\\
IrCNN~\cite{zhang2017IRCNN}   &  32.55 & 34.89 & 33.31 & 32.02 &32.82  & 31.70 & 31.84 & 34.53 & 32.43 & 32.34 &  32.40&   32.40 & 32.769\\

Ours-agnostic       & 32.11  & 35.10 &	33.28 &	\textcolor{blue}{32.31} &	33.07 &	31.58 &	31.80 &	\textcolor{blue}{34.67} &	32.48 &	\textcolor{blue}{32.42} &	32.40 &	 \textcolor{blue}{32.50} & 32.812 \\
Ours-specific              & \textcolor{red}{32.61} & \textcolor{red}{35.21} & \textcolor{blue}{33.21} & \textcolor{red}{32.35} & \textcolor{red}{33.33} & \textcolor{red}{31.77} & \textcolor{red}{32.01} & \textcolor{red}{34.69} & 32.74 & \textcolor{red}{32.44} & \textcolor{red}{32.50} & \textcolor{red}{32.52} & \textcolor{red}{32.950}\\

\midrule

& \multicolumn{13}{c}{$\sigma_n = 25$} \\ 
\cmidrule(l){1-14}

BM3D~\cite{Dabov2007BM3D}     & 29.45 & 32.85 & 30.16 & 28.56 & 29.25 & 28.42 & 28.93 & 32.07 & \textcolor{blue}{30.71} & 29.90 & 29.61 & 29.71 & 29.969\\
WNNM~\cite{Gu2014WNN}         & 29.64 & 33.22 & 30.42 & 29.03 & 29.84 & 28.69 & 29.15 & 32.24 & \textcolor{red}{31.24} & 30.03 & 29.76 & 29.82 & 30.257\\
EPLL~\cite{Zoran2011EPLL}     & 29.26 & 32.17 & 30.17 & 28.51 & 29.39 & 28.61 & 28.95 & 31.73 & 28.61 & 29.74 & 29.66 & 29.53 & 29.692\\
MLP~\cite{Burger2012MLP}      & 29.61 & 32.56 & 30.30 & 28.82 & 29.61 & 28.82 & 29.25 & 32.25 & 29.54 & 29.97 & 29.88 & 29.73 & 30.027\\
CSF~\cite{schmidt2014CSF}     & 29.48 & 32.39 & 30.32 & 28.80 & 29.62 & 28.72 & 28.90 & 31.79 & 29.03 & 29.76 & 29.71 & 29.53 & 29.837\\
TNRD~\cite{chen2017TNRD}      & 29.72 & 32.53 & 30.57 & 29.02 & 29.85 & 28.88 & 29.18 & 32.00 & 29.41 & 29.91 & 29.87 & 29.71 & 30.055\\
DnCNNS~\cite{zhang2017DnCNN}  & \textcolor{blue}{30.18} & 33.06 & 30.87 & 29.41 & 30.28 & \textcolor{blue}{29.13} & \textcolor{blue}{29.43} & 32.44 & 30.00 & 30.21 & \textcolor{blue}{30.10} & 30.12 & 30.436\\
DnCNNB~\cite{zhang2017DnCNN}  & 29.94 & 33.05 & 30.84 & 29.34 & 30.25 & 29.09 & 29.35 & 32.42 & 29.69 & 30.20 & 30.09 & 30.10 & 30.362\\
IrCNN~\cite{zhang2017IRCNN}   & 30.08  & 33.06  & 30.88&  29.27 &  30.09 & 29.12  & 29.47 & 32.43  & 29.92 &  30.17 &  30.04 & 30.08 & 30.384\\

Ours-agnostic   &	29.87 &	\textcolor{blue}{33.34} &	\textcolor{red}{30.94} &	\textcolor{blue}{29.68} &	\textcolor{blue}{30.39} &	29.08 &	29.38 &	\textcolor{blue}{32.65} &	30.17 &	\textcolor{blue}{30.27} &	30.08 &	\textcolor{blue}{30.20} &	\textcolor{blue}{30.505}\\

Ours-specific     & \textcolor{red}{30.26} & \textcolor{red}{33.44} & \textcolor{blue}{30.87} & \textcolor{red}{29.77} & \textcolor{red}{30.62} & \textcolor{red}{29.23} & \textcolor{red}{29.61} & \textcolor{red}{32.66} & 30.29 & \textcolor{red}{30.30} & \textcolor{red}{30.18} & \textcolor{red}{30.24} & \textcolor{red}{30.624}\\
\midrule

& \multicolumn{13}{c}{$\sigma_n = 50$} \\ 
\cmidrule(l){1-14} 
 
BM3D~\cite{Dabov2007BM3D}     &	26.13 & 29.69 & 26.68 & 25.04 & 25.82 & 25.10 & 25.90 & 29.05 & \textcolor{blue}{27.22} & 26.78 & 26.81 & 26.46 & 26.722\\
WNNM~\cite{Gu2014WNN}         &	26.45 & 30.33 & 26.95 & 25.44 & 26.32 & 25.42 & 26.14 & 29.25 & \textcolor{red}{27.79} & 26.97 & 26.94 & 26.64 & 27.052\\
EPLL~\cite{Zoran2011EPLL}     & 26.10 & 29.12 & 26.80 & 25.12 & 25.94 & 25.31 & 25.95 & 28.68 & 24.83 & 26.74 & 26.79 & 26.30 & 26.471\\
MLP~\cite{Burger2012MLP}      & 26.37 & 29.64 & 26.68 & 25.43 & 26.26 & 25.56 & 26.12 & 29.32 & 25.24 & 27.03 & 27.06 & 26.67 & 26.783\\
TNRD~\cite{chen2017TNRD}      & 26.62 & 29.48 & 27.10 & 25.42 & 26.31 & 25.59 & 26.16 & 28.93 & 25.70 & 26.94 & 26.98 & 26.50 & 26.812\\
DnCNNS~\cite{zhang2017DnCNN}  & 27.03 & 30.00 & 27.32 & 25.70 & 26.78 & 25.87 & 26.48 & 29.39 & 26.22 & 27.20 & 27.24 & 26.90 & 27.178\\
DnCNNB~\cite{zhang2017DnCNN}  & 27.03 & 30.02 & 27.39 & 25.72 & 26.83 & \textcolor{blue}{25.89} & 26.48 & 29.38 & 26.38 & 27.23 & 27.23 & 26.91 & 27.206\\ 
IrCNN~\cite{zhang2017IRCNN}  &  26.88 & 29.96& 27.33 & 25.57 & 26.61 &  25.89 & 26.55& 29.40& 26.24 &27.17& 27.17  &26.88 & 27.136\\

Ours-agnostic   &	\textcolor{blue}{27.03} &	\textcolor{blue}{30.48} &	\textcolor{blue}{27.57} &	\textcolor{blue}{26.01} &	\textcolor{blue}{27.03} &	25.84 &	\textcolor{blue}{26.53} &	\textcolor{blue}{29.77} &	26.89 &	\textcolor{blue}{27.28} &	\textcolor{blue}{27.29} &	\textcolor{blue}{27.06} &	\textcolor{blue}{27.398}\\

Ours-specific             & \textcolor{red}{27.25} & \textcolor{red}{30.70} & \textcolor{red}{27.54} & \textcolor{red}{26.05} & \textcolor{red}{27.21} & \textcolor{red}{26.06} & \textcolor{red}{26.53} & \textcolor{red}{29.65} & 26.62 & \textcolor{red}{27.36} & \textcolor{red}{27.26} & \textcolor{red}{27.24} & \textcolor{red}{27.457}\\
 \midrule

& \multicolumn{13}{c}{$\sigma_n = 70$} \\ 
\cmidrule(l){1-14}
BM3D~\cite{Dabov2007BM3D}     & 24.62 & 27.91 &  25.07 &  23.56  & 24.24 &  23.75 & 24.49 & 27.57 & \textcolor{blue}{25.47} & 25.40 & 25.56 & 25.00 & 25.221\\
WNNM~\cite{Gu2014WNN}         & 24.86 & \textcolor{blue}{28.59} &  25.25 &  23.78  & 24.62 &  24.00 & 24.64 & 27.85 & \textcolor{red}{26.17} & 25.58 & 25.68 & 25.18 & 25.517\\
EPLL~\cite{Zoran2011EPLL}     & 24.60 & 27.32 &  25.03 &  23.52  & 24.19 &  23.72 & 24.44 & 27.11 & 23.20 & 25.27 & 25.50 & 24.80 & 24.891\\
DnCNNS~\cite{zhang2017DnCNN}  & \textcolor{blue}{25.37} & 28.22 &  \textcolor{blue}{25.50} &  \textcolor{blue}{23.97}  & \textcolor{blue}{25.10} &  \textcolor{blue}{24.34} & \textcolor{blue}{24.98} & \textcolor{blue}{27.85} & 23.97 & \textcolor{blue}{25.76} & \textcolor{blue}{25.91} & \textcolor{blue}{25.31} & \textcolor{blue}{25.523}\\
Ours-specific           &  \textcolor{red}{25.83} &  \textcolor{red}{29.19} &   \textcolor{red}{25.90} &   \textcolor{red}{24.28}  &  \textcolor{red}{25.66} &   \textcolor{red}{24.59} &  \textcolor{red}{25.12} &  \textcolor{red}{28.25} &  25.06 &  \textcolor{red}{26.00} &  \textcolor{red}{26.02} &  \textcolor{red}{25.78} &  \textcolor{red}{25.974}\\
\bottomrule
\end{tabular}
\label{table:classical_images}
\end{center}
\vspace*{-5mm}
\end{table*}

\subsubsection{Number of modules}
We show the effect of the number of modules on denoising results. As mentioned earlier, each module $\textbf{M}$ consists of six convolution layers, by increasing the number of modules, we are making our network deeper. In this settings, all parameters are constant, except the number of modules as shown in Table~\ref{table:No_of_cnn_modules}. It is clear from the results that making the network deeper increase the average PSNR.  However, since fast restoration is desired, we prefer a small network of three modules \ie $\textbf{M}=3$, which achieves better performance than other methods.

\subsubsection{Kernel dilation and number of layers}
It has been shown that the performance of some networks can be improved either by increasing the depth of the network or by using large convolution filter size to capture the context information~\cite{zhang2017IRCNN,zhang2017DnCNN}. This helps the restoration of noisy structures in the image. The usage of traditional $3\times3$ filters is popular in deeper networks. However, using dilated filters there is a tradeoff between the number of layers and the size of the dilated filters without effecting denoising results. In Table~\ref{table:layers_kernel_dilation_patchsize}, we present three experimental settings to show the tradeoff between the dilated filter size and the depth of network. In the first experiment as shown in the first column of Table~\ref{table:layers_kernel_dilation_patchsize}, we use a traditional filter of size $3\times3$ and depth of 18 to cover the receptive field of training patch of size $40\time40$. In the next experiment, we keep the size of the filter same but enlarge the filter using a dilation factor of two. This increases the size of the filter to $5\times5$ but having nine non-zero entries it can be interpreted as a sparse filter. Therefore, the receptive field of the training patch can now be covered by nine non-linear mapping layers, contrary to the 18-layers depth per module. Similarly, by expanding the filter by a dilation of three would result in the depth of each module to be six. As in Table~\ref{table:layers_kernel_dilation_patchsize}, all three trained models result in similar denoising performance, with the obvious advantage of the shallow network being the most efficient. The number of parameters reduced from 1954k to 663k, similarly, the memory usage for one input patch is reduced to 22MB to 6.5MB.

\begin{table*}
\caption{Performance comparison between our method and existing algorithms on the grayscale version of the BSD68 dataset~\cite{Martin2001BSD}. The missing denoising results, indicated by ``-'', occurs when the method is not trained to deal with the input noisy images.}
\centering
\begin{tabular}{c|c|c|c|c|c|c|c|c|c|c}
\hline\hline
 Noise & \multicolumn{10}{c}{Methods} \\ 
 Level& BM3D    & WNNM  	& EPLL	  & TNRD   & DenoiseNet& DnCNNS  & IrCNN & NLNet	& Ours-Agnostic	& Ours-Specific  \\\hline 
 15    & 31.08   & 31.32 	& 31.19	  & 31.42 & 31.44 & \textcolor{blue}{31.73}   &31.63 & 31.52	&31.68 &\textcolor{red}{31.81} \\ 
 25    & 28.57   & 28.83	& 28.68	  & 28.92 & 29.04 & \textcolor{blue}{29.23}   &29.15 & 29.03 	&29.18 &\textcolor{red}{29.34} \\  
 50    & 25.62   & 25.83	& 25.67	  & 26.01 & 26.06  & 26.23   &26.19 & 26.07	&\textcolor{blue}{26.31} &\textcolor{red}{26.40} \\ 
 70    & 24.44   &    -    	&  24.43  &   -   & -  & \textcolor{blue}{24.90}	&	-  &   -    & -    &\textcolor{red}{25.13} \\ \hline \hline
\end{tabular}
\label{table:BSD68_grayscale}
\vspace*{0.5mm}

\caption{The similarity between the denoised color images and the ground-truth color images of BSD68 dataset for our network and existing algorithms measured by PSNR (in dB) reported for noise levels of $\sigma$=15, 25, and 50.}
\centering
\begin{tabular}{c|c|c|c|c|c|c|c|c}
\hline \hline
Noise  & \multicolumn{8}{c}{Methods} \\ 
Levels & CBM3D~\cite{dabov2007CBM3D}       & MLP~\cite{Burger2012MLP}    & TNRD~\cite{chen2017TNRD}  & DnCNN~\cite{zhang2017DnCNN}  & IrCNN~\cite{zhang2017IRCNN}  & CNLNet~\cite{lefkimmiatis2017NLNet} & Ours-agnostic	& Ours-specific   \\ \hline
 15    &  33.50      & -      & 31.37 & 33.89   & 33.86 & 33.69	& \textcolor{blue}{33.96} & \textcolor{red}{34.12}\\  
 25    &  30.69      & 28.92  & 28.88 & \textcolor{blue}{31.33}   & 31.16 & 30.96   & \textcolor{blue}{31.32} & \textcolor{red}{31.42} \\  
 50    &  27.37      & 26.00  & 25.94 & 27.97   & 27.86 & 27.64	& \textcolor{blue}{28.05} & \textcolor{red}{28.19}\\ \hline  \hline                
\end{tabular}
\label{table:BSD68_color}
\vspace*{-5mm}
\end{table*}

\begin{table}[t!b!]
\caption{Comparisons with state-of-the-art methods on BSD68 with $\sigma_n=50$, and BSD100 with $\sigma_n=25$. The results of \cite{bae2017beyond} and \cite{jiao2017formresnet} are taken from their respective papers.}
\centering
\begin{tabular}{|l|l|l|l|l|}\hline
				& Jiao \cite{jiao2017formresnet}	&Bae~\cite{bae2017beyond}	&DnCNN~\cite{zhang2017DnCNN} 		&Ours 	 \\\hline 
Kernel Size		& 3x3   	    &3x3 	    & 3x3  	    &3x3   \\ \hline   
Patch Size		& 40x40  	    &40x40 		& 40x40     &40x40 \\ \hline
Channels		& 64  	        &320	    & 64 		&64    	 \\ \hline
Training		& BSD400 	    &BSD400  	& BSD400    &BSD400  \\ 
data			& 		 	    &+Urban100 	&       	&    	 \\ \hline
BatchNorm		& Yes  	        &Yes 	    &Yes   		&No    	 \\ \hline
Conv. layers	& 20  	        &20	       	&17		    &19    	 \\ \hline
No. parameters	& 671k  	    &16629k	    &566k 	    &630k    \\ \hline
BSD68			& - 	        &26.35dB    &26.23dB    &\textbf{26.40}dB \\\hline
BSD100			& 29.08dB 	    &-	       	&29.05dB	& \textbf{29.25}dB\\ \hline
\end{tabular}
\label{table:CNN_comparisons}
\vspace*{-5mm}
\end{table}


\subsubsection{Network structure Analysis} 
In Table~\ref{table:feature_contributions}, we show the performance on BSD68 dataset when adding  different features including a kernel dilation of three across all convolution layers, identity skip connection, or boosting via geometric transformation to the DnCNN baseline which is reported in the first row.
The improvement over DnCNN is observed with the introduction of identity skip connections. Applying a dilation of three over 17 or 19 convolutional layers of DnCNN (row 2) does not appear to be effective. However, using dilated convolution in a short chain of six layers, such as row 3, improves the performance further. In Table~\ref{table:feature_contributions}, PSNR is $29.32$ dB without boosting and $29.34$ dB (last row) if we average the output from eight transformed images.


\subsection{Grayscale Image Denoising}

In this section, first we demonstrate how our method performs on classical images and then report results on the BSD68 dataset.


\subsubsection{Classical Images}
For completeness, we compare our algorithm to several state-of-the-art denoising methods using grayscale classical images shown in Figure~\ref{fig:classical_grayscale_image} and reported in Table~\ref{table:classical_images}.

In Table~\ref{table:classical_images}, we present the average PSNR for the denoised images. Our network is the best performer for almost all classical images except \enquote{Barbara}. The reason for this may be the repetitive structures in the mentioned image, which makes it easy for BM3D~\cite{Dabov2007BM3D} and WNNM~\cite{Gu2014WNN} to find and employ patches with great similarity to the noisy input, hence providing better results. 

Subsequently, we depict an example from the classical images. The visual quality of our recovered images, as shown in Figure~\ref{fig:classical_grayscale_image}, is better than all others. This also illustrates that our network restores aesthetically pleasing textures. Small and noticeable features restored by our network include the sharpness and the clarity of the subtle textures around the fore and hind wings, mouth, and antennas of the butterfly. 
Furthermore, a magnified view of the results in Figures~\ref{fig:classical_grayscale_image} for methods such as~\cite{Dabov2007BM3D,zhang2017IRCNN,lefkimmiatis2017NLNet} shows artifacts and failures in the smooth areas. Our CNN network also outperforms~\cite{zhang2017IRCNN,lefkimmiatis2017NLNet,zhang2017DnCNN}, which are trained using deep neural networks.

\subsubsection{BSD68 Dataset}

We present the average PSNR scores for the estimated denoised images in Table~\ref{table:BSD68_grayscale}. The IRCNN~\cite{zhang2017IRCNN} and DnCNN~\cite{zhang2017DnCNN} network structures are similar, hence produce nearly similar results. On the other hand, our method reconstructs the images accurately, achieving higher PSNR then completing methods on all four levels of noise. Furthermore, the difference in PSNR between our method and the state-of-the-art techniques at the higher noise levels.

For a comprehensive evaluation, we demonstrate the visual results on a selected grayscale image from BSD68~\cite{roth2009fields} dataset in Figure~\ref{fig:BSD68_grayscale_image}. In our results, the image details are more similar to the ground-truth details, and our quantitative results are numerically higher than the others. Our method outperforms the second best method by several orders of magnitude (PSNR is computed in the logarithmic scale). Also, note that the denoising results of other CNN based algorithms are comparable to each other. 

In Table~\ref{table:CNN_comparisons}, we compare our method with Bae~\etal~\cite{bae2017beyond}, Jiao~\etal~\cite{jiao2017formresnet} and DnCNN \cite{zhang2017DnCNN} in terms of the training data, model capacity and denoising performance. Our method outperforms the alternatives with a model capacity comparable to DnCNN \cite{zhang2017DnCNN} and Jiao \etal \cite{jiao2017formresnet}, and much lower than Bae \etal \cite{bae2017beyond}.

\subsection{Color Image Denoising}

For noisy color images, we train our network with the noisy RGB input patches of size 40$\times$40 with the corresponding clean ground-truth patches. We only modify the first and last convolution layer of the grayscale network to input and output three channels instead of one channel, keeping all other parameters same as the grayscale network. 

We present the quantitative results in Table~\ref{table:BSD68_color} and qualitative results in Figures~\ref{fig:color_results_sig15} and \ref{fig:color_results_sig25} against benchmark methods including the latest CNN based state-of-the-art color image denoising techniques \cite{zhang2017DnCNN,zhang2017IRCNN,Dabov2007BM3D}. It can be observed that our algorithm attains an improved average PSNR on all three different noise levels for the color version of BSD68 dataset \cite{roth2009fields}. As shown, our method restores true colors closer to their authentic values while others fail and induce false colorizations in certain image regions. Furthermore, a close look reveals that our network reproduces the local texture with much less artifacts and sufficiently sharp details.


\vspace*{-3mm}
\subsection{Darmstadt Noise Dataset: Real-world images}

\begin{figure*}
\begin{center}
\begin{tabular}{c@{ } c@{ } c@{ } c}
\includegraphics[width=.22\textwidth]{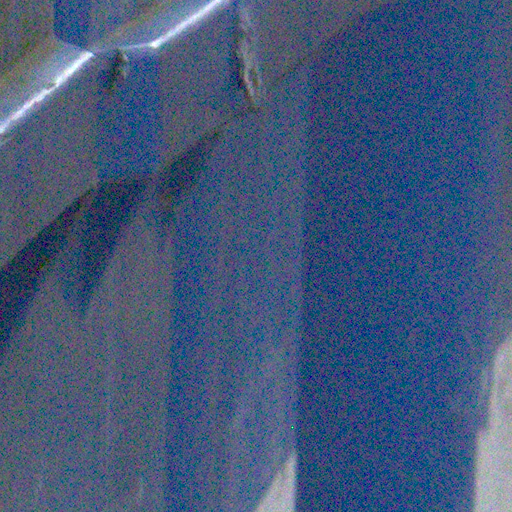}&
\includegraphics[width=.22\textwidth]{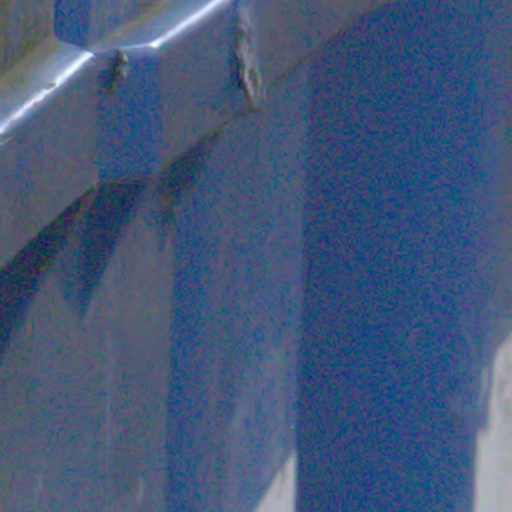}&
\includegraphics[width=.22\textwidth]{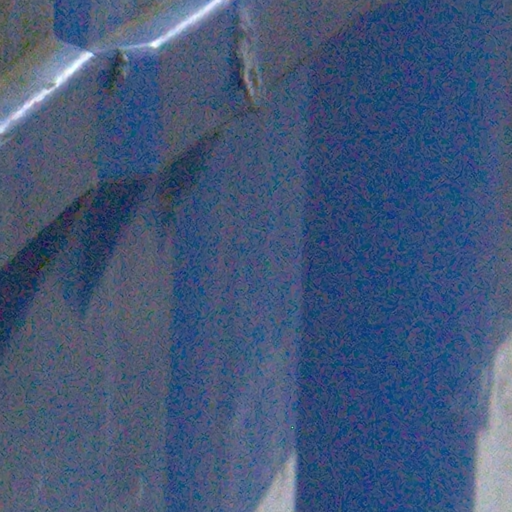}&
\includegraphics[width=.22\textwidth]{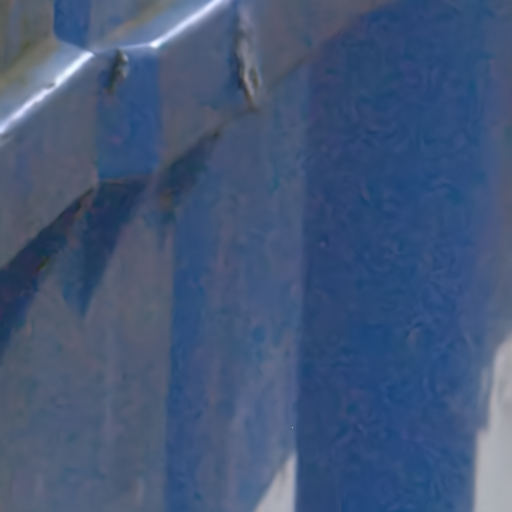}\\
\includegraphics[width=.22\textwidth]{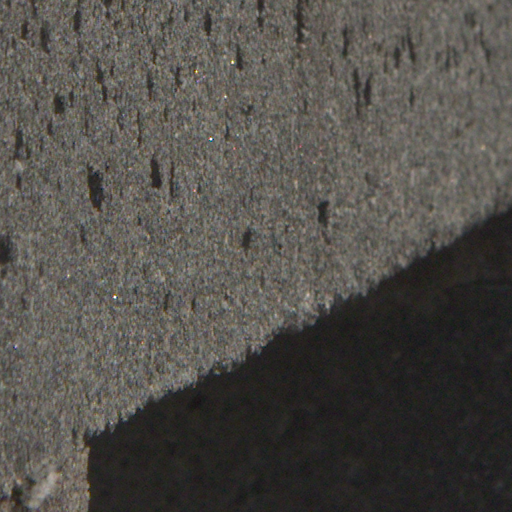}&
\includegraphics[width=.22\textwidth]{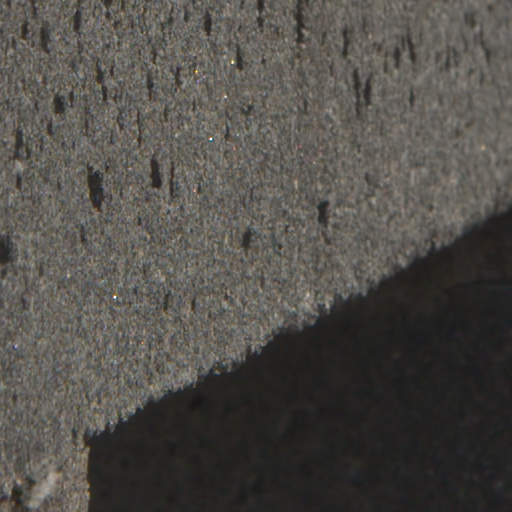}&
\includegraphics[width=.22\textwidth]{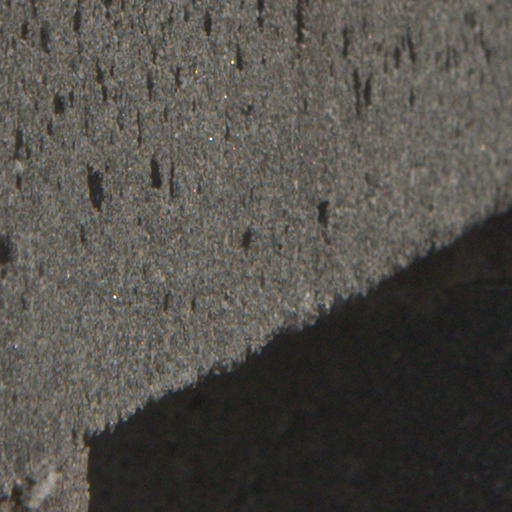}&
\includegraphics[width=.22\textwidth]{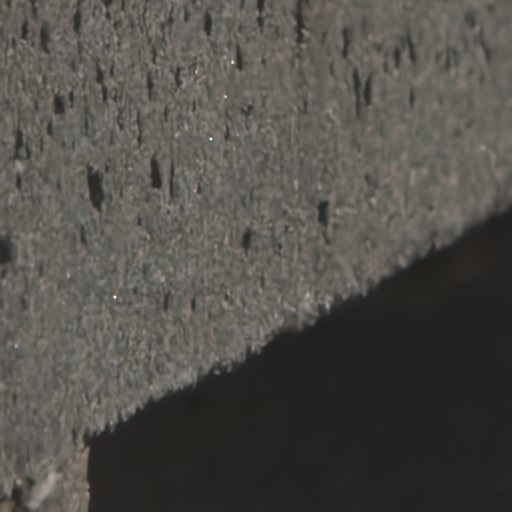}\\
\includegraphics[width=.22\textwidth]{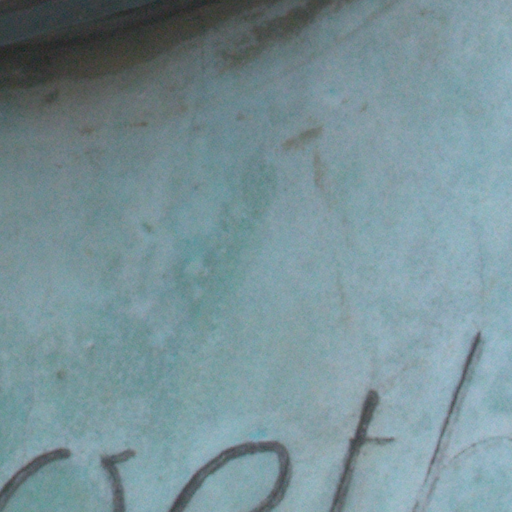}&
\includegraphics[width=.22\textwidth]{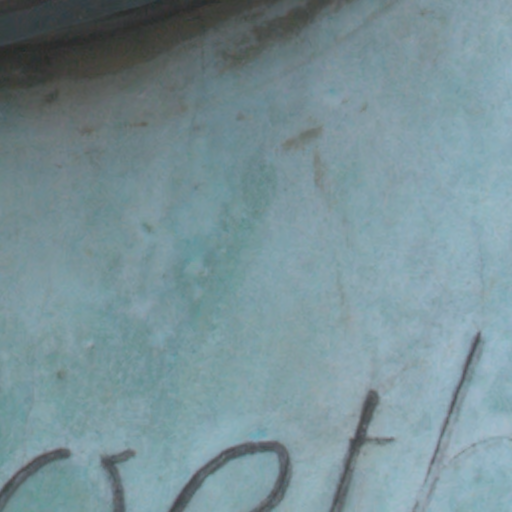}&
\includegraphics[width=.22\textwidth]{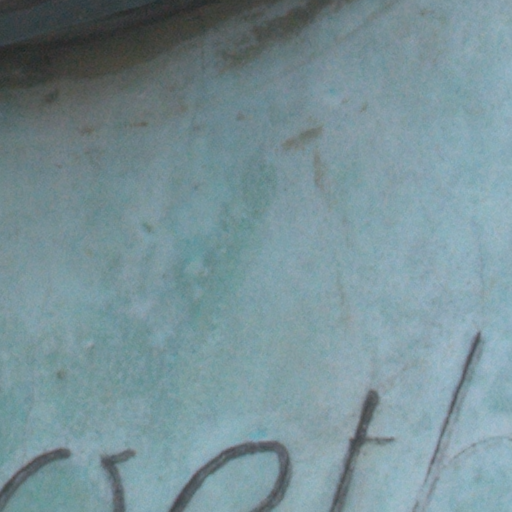}&
\includegraphics[width=.22\textwidth]{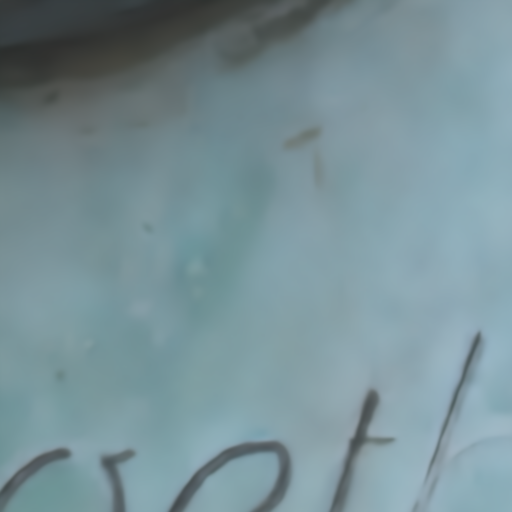}\\
Noisy  & CBM3D & DnCNN & Ours \\
\end{tabular}
\end{center}
\vspace*{-3mm}
\caption{Comparison of our method against the state-of-the-art algorithms on real images containing Gaussian noise from Darmstadt Noise Dataset (DND) benchmark for different denoising algorithms \cite{plotz2017benchmarking}. Difference can be better viewed in magnified view.}
\label{fig:DnD}
\vspace*{-5mm}
\end{figure*}

So far, state-of-the-art denoising methods, such as FormResNet \cite{jiao2017formresnet}, DnCNN \cite{zhang2017DnCNN}, IrCNN \cite{zhang2017IRCNN}  and BM3D \cite{Dabov2007BM3D} \etc have normally been evaluated on classical images and the BSD68 dataset. Recently, \cite{plotz2017benchmarking} proposed the Darmstadt Noise Dataset (DND) benchmark for denoising algorithms which consists of 50 images. The dataset is composed of images with interesting and challenging structures. The images are converted to sRGB and gamma correction is applied. The size of each image is in Megapixels; therefore, each image is cropped at 20 locations and each composed of 512 $\times$ 512 pixels yielding 1000 test crops, and overlap between the images is about 10$\%$. Only these test images are provided, there are no images for either training or validation. Therefore, we use the same model which is trained on the synthetic BSD68 dataset \cite{roth2009fields}.

The quantitative results in PSNR and SSIM averaged over all the images for real-world DnD is presented in Table~\ref{table:DnD_benchmark}. It can be observed that our method is the best performer followed by BM3D. Previously, the classical method BM3D is considered to be outperformed by most of the state-of-the-art algorithms on synthetic datasets; however, this is not the case when using the real-world Darmstadt Noise Dataset. It is to be noted that our method does not require to know the noise level in advance unlike BM3D and other state-of-the-art techniques. Furthermore, we visually compare our method with a few recent algorithms as shown on several samples from \cite{plotz2017benchmarking} in Figure~\ref{fig:DnD}\footnote{PSNR for individual images are not available as \cite{plotz2017benchmarking}'s system only provide average PSNR}. It can be observed that both CBM3D~\cite{Dabov2007BM3D}, as well as DnCNN~\cite{zhang2017DnCNN}, are unable to remove the noise from the images. On the other hand, it can be seen that our method eliminates the noise and preserve the structures.

\vspace*{-3mm}
\section{Conclusions}
\label{sec:Conclusion}
To sum up, we employ residual learning and identity mapping to predict the denoised image using a five-module and five-layer deep network of 26 weight layers with dilated convolutional filters without batch normalization. Our choice of network is based on the ablation studies performed in the experimental section of this paper. 

This is the first modular framework to predict the denoised output without any dependency on the pre- or post-processing. Our proposed network removes the potentially authentic image structures while allowing the noisy observations to go through its layers, and learns the noise patterns to estimate the clean image. 

In synthetic images case, we have provided ample examples and have shown that our network outperforms classical state-of-the-art denoising algorithms that are intended for use on natural images. Furthermore, we have compared against the current convolutional neural networks both visually and numerically. Our network gain is about $\textbf{0.1}$dB on BSD68 dataset~\cite{Martin2001BSD} and results are visually pleasing.

On real images of Darmstadt Noise Dataset (DND) \cite{plotz2017benchmarking}, we have shown that our method provides visually pleasing results and a gain of about $\textbf{1.43}$dB of PSNR. The real images appear less grainy after passing through our proposed network and preserving fine image structures. Furthermore, competitive denoising algorithms require information about the noise in the image while on the contrary, our network does not require any information about the noise present in the images. 

In future, we aim to generalize our denoising network to other image restoration and enhancement tasks such as deblurring, color correction, JPEG artifact removal, rain removal, dehazing and super-resolution \etc

For the time being our approach is only applicable to Gaussian noise removal. However, we would like to train our model with different noise types such as Poisson, astronomical \etc and examine its performance on these specific noise types. It should be noted that other state-of-the-art methods, for example, BM3D \cite{Dabov2007BM3D}, WNNM \cite{Gu2014WNN} are only applicable to Gaussian noise and may not be readily adapted to handle different noise types.

All CNN approaches performance on images with regular, and repeating structures such as ``Barbara'' is relatively less in terms of PSNR compared to classical denoising methods. This phenomenon is due to the design of traditional denoising methods to exploit the regular and repeating structures. To overcome this issue, either block-matching scheme can be incorporated into our CNN approach or relying on consolidating the outcome of various denoising algorithms with our CNN approach.

\begin{table*}
\caption{Mean PSNR and SSIM of the denoising methods evaluated on the real images dataset by \cite{plotz2017benchmarking}. 
}
\centering
\begin{tabular}{|c|c|c|c|c|c|c|c|c|}\hline
& \multicolumn{8}{c|}{Methods}\\ \cline{2-9}
Metrics &WNNM~\cite{Gu2014WNN}   &  EPLL~\cite{Zoran2011EPLL}   &  BM3D~\cite{Dabov2007BM3D}   & MLP~\cite{Burger2012MLP}    & DenoiseNet\cite{DenoiseNet2017} &TNRD~\cite{chen2017TNRD}    & DnCNN~\cite{zhang2017DnCNN}  & Ours-agnostic \\ \hline 
PSNR   &34.44  &  33.51  &  \textcolor{blue}{34.61}  & 34.14  &35.08 & 29.92   & 32.43  & \textcolor{red}{36.04} \\
SSIM   &\textcolor{blue}{0.8646} &  0.8244 &  0.8507 & 0.8331 & 0.8680& 0.8306  & 0.7900 &\textcolor{red}{0.9136} \\ \hline
\end{tabular}
\label{table:DnD_benchmark}
\vspace*{-5mm}
\end{table*}



\ifCLASSOPTIONcaptionsoff
  \newpage
\fi
\bibliographystyle{IEEEtran}
\bibliography{references.bib}

\begin{thebibliography}{10}
\providecommand{\url}[1]{#1}
\csname url@samestyle\endcsname
\providecommand{\newblock}{\relax}
\providecommand{\bibinfo}[2]{#2}
\providecommand{\BIBentrySTDinterwordspacing}{\spaceskip=0pt\relax}
\providecommand{\BIBentryALTinterwordstretchfactor}{4}
\providecommand{\BIBentryALTinterwordspacing}{\spaceskip=\fontdimen2\font plus
\BIBentryALTinterwordstretchfactor\fontdimen3\font minus
  \fontdimen4\font\relax}
\providecommand{\BIBforeignlanguage}[2]{{%
\expandafter\ifx\csname l@#1\endcsname\relax
\typeout{** WARNING: IEEEtran.bst: No hyphenation pattern has been}%
\typeout{** loaded for the language `#1'. Using the pattern for}%
\typeout{** the default language instead.}%
\else
\language=\csname l@#1\endcsname
\fi
#2}}
\providecommand{\BIBdecl}{\relax}
\BIBdecl

\bibitem{zhang2017IRCNN}
K.~Zhang, W.~Zuo, S.~Gu, and L.~Zhang, ``Learning deep cnn denoiser prior for
  image restoration,'' \emph{CVPR}, 2017.

\bibitem{zhang2017DnCNN}
K.~Zhang, W.~Zuo, Y.~Chen, D.~Meng, and L.~Zhang, ``Beyond a gaussian denoiser:
  Residual learning of deep cnn for image denoising,'' \emph{TIP}, 2017.

\bibitem{Liu2018TMMObject}
M.~Liu, H.~Liu, and C.~Chen, ``Robust 3d action recognition through sampling
  local appearances and global distributions,'' \emph{TMM}, 2018.

\bibitem{Liu2016FaceTMM}
L.~Liu, C.~Xiong, H.~Zhang, Z.~Niu, M.~Wang, and S.~Yan, ``Deep aging face
  verification with large gaps,'' \emph{TMM}, 2016.

\bibitem{Liu2018InpaintingTMM}
J.~Liu, S.~Yang, Y.~Fang, and Z.~Guo, ``Structure-guided image inpainting using
  homography transformation,'' \emph{TMM}, 2018.

\bibitem{Galteri2019ArtifactTMM}
L.~Galteri, L.~Seidenari, M.~Bertini, and A.~D. Bimbo, ``Deep universal
  generative adversarial compression artifact removal,'' \emph{TMM}, 2019.

\bibitem{Yu2014Deblurring}
X.~Yu, F.~Xu, S.~Zhang, and L.~Zhang, ``Efficient patch-wise non-uniform
  deblurring for a single image,'' \emph{TMM}, 2014.

\bibitem{Xie2015SRDenoiseTMM}
J.~Xie, R.~S. Feris, S.~Yu, and M.~Sun, ``Joint super resolution and denoising
  from a single depth image,'' \emph{TMM}, 2015.

\bibitem{Yang2015learningTMM}
X.~Yang, T.~Zhang, and C.~Xu, ``Cross-domain feature learning in multimedia,''
  \emph{TMM}, 2015.

\bibitem{Buades2005NLM}
A.~Buades, B.~Coll, and J.-M. Morel, ``A non-local algorithm for image
  denoising,'' in \emph{CVPR}, 2005, pp. 60--65.

\bibitem{Dabov2007BM3D}
K.~Dabov, A.~F., V.~Katkovnik, and K.~Egiazarian, ``Image denoising by sparse
  3-{D} transform-domain collaborative filtering,'' 2007, pp. 2080--2095.

\bibitem{Gu2014WNN}
S.~Gu, L.~Zhang, W.~Zuo, and X.~Feng, ``Weighted nuclear norm minimization with
  application to image denoising,'' in \emph{CVPR}, 2014, pp. 2862--2869.

\bibitem{peng2012rasl}
Y.~Peng, A.~Ganesh, J.~Wright, W.~Xu, and Y.~Ma, ``Rasl: Robust alignment by
  sparse and low-rank decomposition for linearly correlated images,''
  \emph{TPAMI}, pp. 2233--2246, 2012.

\bibitem{Huang2014SparseTMM}
D.~Huang, L.~Kang, Y.~F. Wang, and C.~Lin, ``Self-learning based image
  decomposition with applications to single image denoising,'' \emph{TMM},
  2014.

\bibitem{xu2007Iterative}
J.~Xu and S.~Osher, ``Iterative regularization and nonlinear inverse scale
  space applied to wavelet-based denoising,'' \emph{TIP}, pp. 534--544, 2007.

\bibitem{weiss2007makes}
Y.~Weiss and W.~T. Freeman, ``What makes a good model of natural images?'' in
  \emph{CVPR}, 2007, pp. 1--8.

\bibitem{roth2009fields}
S.~Roth and M.~J. Black, ``Fields of experts,'' \emph{IJCV}, 2009.

\bibitem{Yue2014CID}
H.~Yue, X.~Sun, J.~Yang, and F.~Wu, ``Cid: Combined image denoising in spatial
  and frequency domains using web images,'' in \emph{CVPR}, 2014.

\bibitem{anwar2017category}
S.~Anwar, F.~Porikli, and C.~P. Huynh, ``Category-specific object image
  denoising,'' \emph{TIP}, pp. 5506--5518, 2017.

\bibitem{luo2015adaptive}
E.~Luo, S.~H. Chan, and T.~Q. Nguyen, ``Adaptive image denoising by targeted
  databases,'' \emph{TIP}, pp. 2167--2181, 2015.

\bibitem{lefkimmiatis2017NLNet}
S.~Lefkimmiatis, ``Non-local color image denoising with convolutional neural
  networks,'' \emph{CVPR}, 2016.

\bibitem{Foi2007SADCT}
A.~Foi, V.~Katkovnik, and K.~Egiazarian, ``Pointwise shape-adaptive {DCT} for
  high-quality denoising and deblocking of grayscale and color images,''
  \emph{TIP}, pp. 1395--1411, 2007.

\bibitem{Lebrun2013NLB}
M.~Lebrun, A.~Buades, and J.-M. Morel, ``A nonlocal bayesian image denoising
  algorithm,'' \emph{SIAM Journal on Imaging Sciences}, 2013.

\bibitem{Goossens2008INLM}
B.~Goossens, H.~Luong, A.~Pizurica, and W.~Philips, ``An improved non-local
  denoising algorithm,'' in \emph{IP}, 2008, p. 143.

\bibitem{Levin2011Bounds}
A.~Levin and B.~Nadler, ``Natural image denoising: Optimality and inherent
  bounds,'' in \emph{CVPR}, 2011, pp. 2833--2840.

\bibitem{Chatterjee2010IDD}
P.~Chatterjee and P.~Milanfar, ``Is denoising dead?'' \emph{TIP}, 2010.

\bibitem{chan2014monte}
S.~H. Chan, T.~Zickler, and Y.~M. Lu, ``Monte carlo non-local means: Random
  sampling for large-scale image filtering,'' \emph{TIP}.

\bibitem{Elad2009ERD}
M.~Elad and D.~Datsenko, ``Example-based regularization deployed to
  super-resolution reconstruction of a single image,'' \emph{Comput. J.}, 2009.

\bibitem{Mairal2009NLSM}
J.~Mairal, F.~Bach, J.~Ponce, G.~Sapiro, and A.~Zisserman, ``Non-local sparse
  models for image restoration,'' in \emph{ICCV}, 2009.

\bibitem{Dong2011CSR}
W.~Dong, X.~Li, D.~Zhang, and G.~Shi, ``Sparsity-based image denoising via
  dictionary learning and structural clustering,'' in \emph{CVPR}, 2011.

\bibitem{Zha2017Group}
Q.~W. Y.~B. Zhiyuan~Zha, Xinggan~Zhang and L.~Tang, ``Group sparsity residual
  constraint for image denoising,'' in \emph{preprint}, 2017.

\bibitem{Zoran2011EPLL}
D.~Zoran and Y.~Weiss, ``From learning models of natural image patches to whole
  image restoration,'' in \emph{ICCV}, 2011, pp. 479--486.

\bibitem{Chen2015External}
L.~Z. F.~Chen and H.~Yu, ``{External Patch Prior Guided Internal Clustering for
  Image Denoising},'' in \emph{ICCV}, 2015.

\bibitem{Xu2015PG-GMM}
J.~Xu, L.~Zhang, W.~Zuo, D.~Zhang, and X.~Feng, ``{Patch Group Based Nonlocal
  Self-Similarity Prior Learning for Image Denoising},'' in \emph{ICCV}, 2015,
  pp. 1211--1218.

\bibitem{Xu2015PGPD}
L.~Xu, L.~Zhang, W.~Zuo, D.~Zhang, and X.~Feng, ``Patch group based nonlocal
  self-similarity prior learning for image denoising,'' 2015.

\bibitem{chen2015PCLR}
F.~Chen, L.~Zhang, and H.~Yu, ``External patch prior guided internal clustering
  for image denoising,'' 2015.

\bibitem{Yue2015CID}
H.~Yue, X.~Sun, J.~Yang, and F.~Wu, ``Image denoising by exploring external and
  internal correlations,'' \emph{TIP}, pp. 1967--1982, 2015.

\bibitem{schmidt2014CSF}
U.~Schmidt and S.~Roth, ``Shrinkage fields for effective image restoration,''
  in \emph{CVPR}, 2014.

\bibitem{Burger2012MLP}
H.~C. Burger, C.~J. Schuler, and S.~Harmeling, ``Image denoising: Can plain
  neural networks compete with bm3d?'' in \emph{CVPR}, 2012.

\bibitem{chen2017TNRD}
Y.~Chen and T.~Pock, ``Trainable nonlinear reaction diffusion: A flexible
  framework for fast and effective image restoration,'' \emph{TPAMI}, pp.
  1256--1272, 2017.

\bibitem{jiao2017formresnet}
J.~Jiao, W.-C. Tu, S.~He, and R.~W. Lau, ``Formresnet: Formatted residual
  learning for image restoration,'' in \emph{CVPRW}, 2017.

\bibitem{bae2017beyond}
W.~Bae, J.~Yoo, and J.~C. Ye, ``Beyond deep residual learning for image
  restoration: Persistent homology-guided manifold simplification,'' in
  \emph{CVPRW}, 2017.

\bibitem{edelsbrunner2008persistent}
H.~Edelsbrunner and J.~Harer, ``Persistent homology-a survey,''
  \emph{Contemporary mathematics}, pp. 257--282, 2008.

\bibitem{bengio1994vanishing}
Y.~Bengio, P.~Simard, and P.~Frasconi, ``Learning long-term dependencies with
  gradient descent is difficult,'' \emph{TNN}, 1994.

\bibitem{he2016deep}
K.~He, X.~Zhang, S.~Ren, and J.~Sun, ``Deep residual learning for image
  recognition,'' in \emph{CVPR}, 2016, pp. 770--778.

\bibitem{dabov2007CBM3D}
K.~Dabov, A.~Foi, V.~Katkovnik, and K.~Egiazarian, ``Color image denoising via
  sparse 3-{D} collaborative filtering with grouping constraint in
  luminance-chrominance space,'' in \emph{ICIP}, 2007.

\bibitem{He15Residual}
K.~H., X.~Zhang, S.~Ren, and J.~Sun, ``Deep residual learning for image
  recognition,'' \emph{CoRR}, 2015.

\bibitem{He2016IM}
K.~He, X.~Zhang, S.~Ren, and J.~Sun, ``Identity mappings in deep residual
  networks,'' \emph{ECCV}, 2016.

\bibitem{Lin16FPN}
T.~Lin, P.~Doll{\'{a}}r, R.~B. Girshick, K.~He, B.~Hariharan, and S.~J.
  Belongie, ``Feature pyramid networks for object detection,'' \emph{CoRR},
  2016.

\bibitem{kim2016VDSR}
J.~Kim, J.~Kwon~Lee, and K.~Mu~Lee, ``Accurate image super-resolution using
  very deep convolutional networks,'' in \emph{CVPR}, 2016.

\bibitem{Martin2001BSD}
D.~Martin, C.~Fowlkes, D.~Tal, and J.~Malik, ``A database of human segmented
  natural images and its application to evaluating segmentation algorithms and
  measuring ecological statistics,'' in \emph{ICCV}, 2001.

\bibitem{plotz2017benchmarking}
T.~Pl{\"o}tz and S.~Roth, ``Benchmarking denoising algorithms with real
  photographs,'' \emph{arXiv preprint arXiv:1707.01313}, 2017.

\bibitem{KingmaB14}
D.~P. Kingma and J.~Ba, ``Adam: {A} method for stochastic optimization,''
  \emph{CoRR}, 2014.

\bibitem{He15Rectifiers}
K.~He, X.~Zhang, S.~Ren, and J.~Sun, ``Delving deep into rectifiers: Surpassing
  human-level performance on imagenet classification,'' \emph{CoRR}, 2015.

\bibitem{timofte2016seven}
R.~Timofte, R.~Rothe, and L.~Van~Gool, ``Seven ways to improve example-based
  single image super resolution,'' in \emph{CVPR}, 2016.

\bibitem{DenoiseNet2017}
T.~Remez, O.~Litany, R.~Giryes, and A.~M. Bronstein, ``Deep class-aware image
  denoising,'' in \emph{International Conference on Sampling Theory and
  Applications}, 2017.

\end{thebibliography}

\vspace*{-15mm}
\begin{IEEEbiography}[{\includegraphics[width=\textwidth]{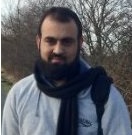}}] {Saeed Anwar}
received Bachelor degree in Computer Systems Engineering with distinction from University of Engineering and Technology (UET), Pakistan, in July 2008, and Master degree in Erasmus Mundus Vision and Robotics (Vibot), Europe, in August 2010 with distinction. During his masters, he carried out his thesis at Toshiba Medical Visualization Systems Europe, Scotland. He has also been a visiting research fellow at Pal Robotics, Barcelona. Since 2014, he is a PhD student at ANU and Data61/CSIRO. He has also been working as a Lecturer and Assistant Professor at NUCES, Pakistan. His major research interests are low-level vision, enhancement, restoration, $\&$ optimization.
\end{IEEEbiography}
\vspace{-15mm}
\begin{IEEEbiography}[{\includegraphics[width=\textwidth]{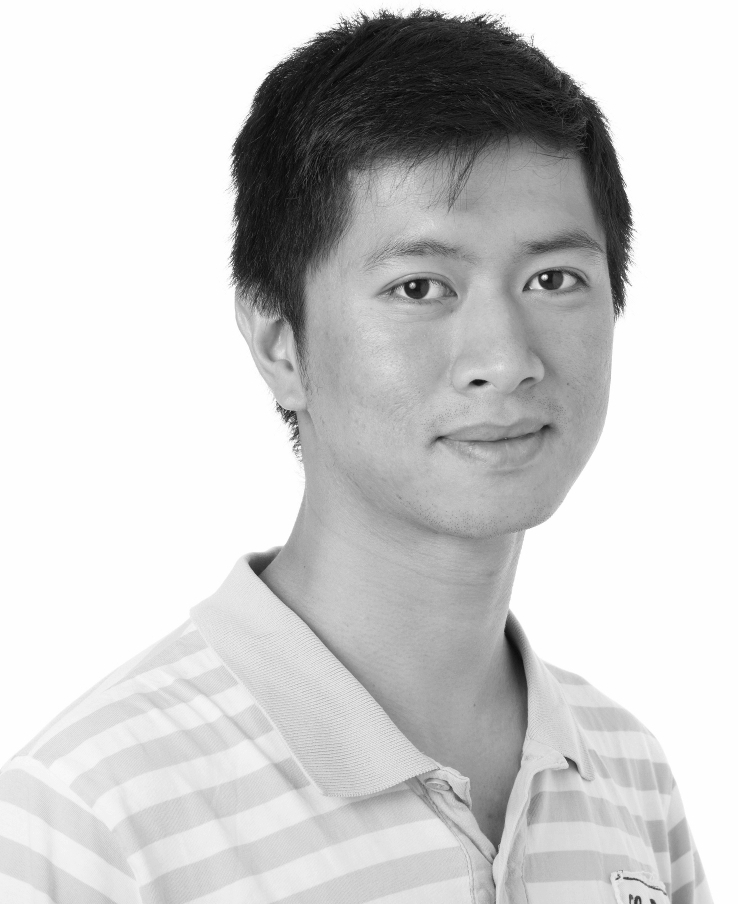}}] {Cong Phuoc Huynh} is a research scientist at Amazon Lab126, and   concurrently an adjunct research fellow at ANU. He has co-authored a book on imaging spectroscopy for scene analysis and over 20 journal articles and conference papers in computer vision and pattern recognition. He is an inventor of eight patents on spectral imaging. He is a co-recipient of a DICTA Best Student's paper Award in 2013. Previously, he was a computer vision researcher at NICTA. He received a B.Sc. degree (Hons) in Computer Science and Software Engineering from the University of Canterbury, New Zealand in 2006, and M.Sc. and Ph.D. degrees in Computer Science from ANU in 2007 and 2012. 
\end{IEEEbiography}
\vspace{-15mm}
\begin{IEEEbiography}[{\includegraphics[width=\textwidth]{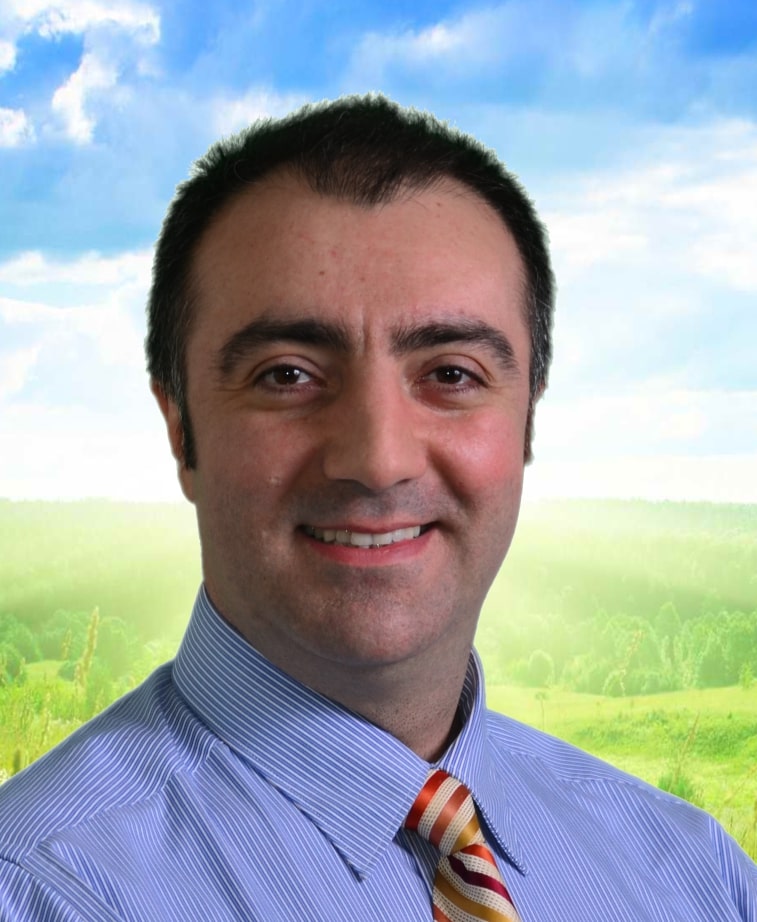}}]{Fatih Porikli} is an IEEE Fellow and a Professor in the Research School of Engineering, ANU. He is also managing the Computer Vision Research Group at Data61/CSIRO. He has received his PhD from New York University in 2002. Previously he served Distinguished Research Scientist at Mitsubishi Electric Research Laboratories. Prof. Porikli is the recipient of the R\&D 100 Scientist of the Year Award in 2006. He won 4 best paper awards at premier IEEE conferences and received 5 other professional prizes. Prof. Porikli authored more than 150 publications and invented 66 patents. He is the co-editor of 2 books. He is serving as the Associate Editor of 5 journals for the past 8 years. He was the General Chair of AVSS 2010 and WACV 2014, and the Program Chair of WACV 2015 and AVSS 2012. His research interests include computer vision, deep learning, manifold learning, online learning, and image enhancement with commercial applications in video surveillance, car navigation, intelligent transportation, satellite, and medical systems.
\end{IEEEbiography}

\end{document}